\documentclass[10pt]{article} 

\usepackage[preprint]{tmlr}

\usepackage{amsmath,amsfonts,bm}

\def\eqref#1{equation~\ref{#1}}

\def\1{\bm{1}}

\DeclareMathAlphabet{\mathsfit}{\encodingdefault}{\sfdefault}{m}{sl}
\SetMathAlphabet{\mathsfit}{bold}{\encodingdefault}{\sfdefault}{bx}{n}

\usepackage{hyperref}
\usepackage{url}

\usepackage{amsmath,amssymb}
\usepackage{booktabs}
\usepackage{multirow}
\usepackage{placeins}
\usepackage{graphicx}

\title{How Far Can INRs Go? Cross-Domain Parameter-efficient INR-Based Semantic Segmentation for Brain MRI}

\author{\name Ziyao Shang \email z27shang@uwaterloo.ca \\
      \addr Department of Systems Design Engineering\\
      University of Waterloo
      \AND
      \name Pouya Sadeghi \email pouya.sadeghi@uwaterloo.ca \\
      \addr David R. Cheriton School of Computer Science\\
      University of Waterloo
      \AND
      \name Letian Jiang \email l236jiang@uwaterloo.ca\\
      \addr David R. Cheriton School of Computer Science \\
      University of Waterloo
      \AND
      \name Alexander Wong \email a28wong@uwaterloo.ca\\
      \addr Department of Systems Design Engineering\\
      University of Waterloo
      \AND
      \name Sirisha Rambhatla \email sirisha.rambhatla@uwaterloo.ca\\
      \addr Department of Management Science and Engineering \\
      University of Waterloo 
      }

\def\month{MM}  
\def\year{YYYY} 
\def\openreview{\url{https://openreview.net/forum?id=XXXX}}

\begin{document}

\maketitle

\begin{abstract}
Biomedical image segmentation is central to medical image analysis, but practical deployment often faces limited annotations, memory constraints, and cross-site distribution shifts. Implicit Neural Representations (INRs) have recently emerged as a lightweight alternative for semantic segmentation, achieving competitive performance with substantially fewer parameters than conventional architectures. However, the mechanisms, scaling behavior, and domain generalization abilities of INR-based segmentation remain insufficiently understood. In this work, we study these questions in the context of cross-domain brain MRI segmentation. We analyze INR-based segmentation across low-parameter regimes, comparing it with conventional pipelines in both in-domain and out-of-domain settings. Surprisingly, we find that INR-based models do not simply improve with increasing parameter budget. Their advantage is most pronounced under low-parameter and limited-augmentation settings, while U-Net-based models benefit more from larger capacity and standard augmentation. We also investigate how INRs encode semantic information in their hidden features and show that complementary segmentation-relevant structure is distributed across multiple INR layers. Building on this insight, we introduce HierINRSeg, a hierarchical INR-based architecture that aggregates multi-layer representations for improved robustness and generalization. Extensive experiments show that HierINRSeg consistently outperforms MetaSeg, a strong recent INR-based segmentation baseline, with an average improvement of 5.6 percentage points in Dice for the in-domain test set and 8.2 percentage points out-of-domain. Overall, our analysis identifies the conditions under which INR-based segmentation is most effective, providing concrete guidance for model selection and future research\footnote{The authors have used large language models to assist with manuscript writing and code development. All content was reviewed and verified by the authors.}. 
\end{abstract}

\section{Introduction}
\label{sec:intro}

Semantic segmentation, where each pixel in an image is categorized into a specific class, is a key dense prediction problem in computer vision, with applications ranging from autonomous navigation~\cite{cordts2016cityscapes} to computer-aided medical diagnosis~\cite{azad2022medical, litjens2017survey}. Fully convolutional networks~\cite{long2015fcn}, U-Net-based models~\cite{ronneberger2015unet}, and transformer-based architectures~\cite{zheng2021setr, xie2021segformer} have produced highly capable models and pipelines for biomedical image segmentation, such as nnUNet~\cite{isensee2021nnu, isensee2024nnunet}, UNeXT~\cite{valanarasu2022unext}, and TransUNet~\cite{chen2021transu}. However, such capability often depends on substantial model capacity, large annotated datasets, carefully engineered augmentations~\cite{isensee2021nnu}, and the implicit assumption that training and testing sets share the same distribution.

\begin{figure}[tb]
    \centering
    \includegraphics[width=0.98\linewidth]{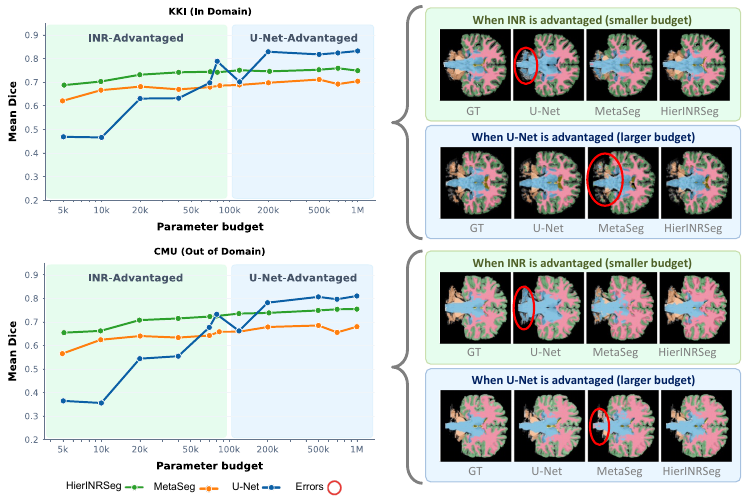}
    \caption{Scaling comparison of U-Net- and INR-based segmentation models under training augmentation and domain shift. Results are shown on KKI, the source (in-domain) site, and CMU, a target (out-of-domain) site. Left: foreground mean Dice across parameter budgets for U-Net, MetaSeg, and our HierINRSeg, demonstrating stronger performance for INR-based models in the low-parameter regime, while U-Net becomes more competitive as model capacity increases. Right: representative segmentation results at lower- and higher-capacity settings, showing a transition from INR-advantaged performance at smaller model sizes to U-Net-advantaged performance at larger model sizes. Red circles highlight typical segmentation errors. The ground truth segmentation is denoted as GT.}
    \label{fig:teaser}
\end{figure}

These requirements are difficult to satisfy in practice, especially in biomedical imaging. Expert annotations are scarce and costly, while deployment in resource-constrained clinical or edge settings often requires models with limited compute and memory~\cite{Wang_2021, ZHANG2024108088}. For brain MRI segmentation, such challenges~\cite{zhang2021quality, ali2022implementation} are compounded by cross-site domain shift, where a network trained at one site degrades on images from a different scanner, protocol, or patient population~\cite{albadawy2018cross}. This poses a question that conventional architectures answer poorly: can we segment brain MR images accurately with very few parameters and without relying on heavy augmentation, while remaining robust when the test distribution drifts? A recent alternative comes from Implicit Neural Representations (INRs), a family of models originally developed for compactly encoding signals.

An INR represents a signal as a continuous function that maps coordinates to values through a small neural network~\cite{sitzmann2020implicit, tancik2020fourier, essakine2025stand}. Initially developed for representing complex visual signals~\cite{mescheder2019occupancy, park2019deepsdf, mildenhall2020nerf}, INRs are highly parameter-efficient, making them suitable under low parameter budgets~\cite{dupont2021coin}. However, by construction, a single INR overfits one signal and does not naturally generalize across a distribution of signals~\cite{vyas2025learningtrans}. A growing body of work removes this barrier through meta-learning or sharing structure across instances:~\cite{tancik2021learned, dupont2022functa, wu2026disentangled}. Such strategies yield INRs possessing consistency across signals, making them usable for downstream prediction, such as dense labeling~\cite{zhi2021ilabel, kohli2021semantic, kundu2022panoptic} and medical image segmentation~\cite{10.1007/978-3-031-16443-9_42, wei2024imedsam, stoltansnisf}. Closest to our setting, MetaSeg~\cite{VyaKus_Fit2025} meta-learns a shared INR that is fit to each test image and feeds the fitted network's hidden features to a lightweight segmentation head, matching conventional segmentation accuracy with a fraction of the parameters and establishing INRs as a way to parameter-efficient brain MRI segmentation.

Despite this promise, INR-based segmentation remains poorly understood in two respects. First, it is unclear \emph{why} it works: information reaches the segmentation head only indirectly through the fitted INR's hidden features, yet the organization of semantic structure within those features has not been examined. Second, it is unclear \emph{when} INRs are preferred: INR-based segmentation has not been systematically compared with conventional CNNs across parameter budgets, augmentation settings, and domain shifts, leaving limited understanding of how their relative performance evolves across different resource regimes. We study both questions using low-parameter cross-site brain MRI segmentation, a setting that captures challenges both in limited model capacity and in cross-domain generalization.

Our study makes three contributions. First, we analyze the layer-wise feature geometry of meta-trained INRs~\cite{VyaKus_Fit2025} and show that segmentation-relevant information is distributed across the hierarchy: early and middle layers retain stable, transferable spatial-anatomical structure, whereas later layers become more label-discriminative and image-specific. Second, guided by this analysis, we introduce \textbf{HierINRSeg}, an INR-based segmentation architecture that hierarchically fuses multi-layer INR features, and show that it consistently outperforms MetaSeg~\cite{VyaKus_Fit2025}, a strong recent INR baseline, in both in-domain and out-of-domain settings, with an improvement of 5.6 Dice percentage in-domain and improvements of 7.2 and 9.2 on the two out-of-domain sites, averaged across all parameter budgets. Third, through a controlled comparison spanning 5K to 1M parameters, with and without augmentation, we characterize how the relative performance of INR- and U-Net-based models changes, as summarized in Figure~\ref{fig:teaser}. Interestingly, increasing model capacity does not consistently benefit INR-based segmentation: their advantage lies in low-parameter and limited-augmentation settings, while U-Net-based models become increasingly competitive at larger capacities and benefit more from data augmentation, a difference we further analyze by tracking augmentation-induced feature changes during INR fitting.

\section{Related work}
\subsection{Implicit Neural Representations} 
Implicit neural representations (INRs) are continuous functions
parameterized by neural networks trained to represent signals~\cite{sitzmann2020implicit, essakine2025stand}. INRs have been commonly used in scene representation/neuro-rendering~\cite{M_ller_2022, mildenhall2020nerf, yu2021pixelnerf}, image compression/super-resolution~\cite{10.1007/978, Jiang_2025}, and scientific simulation/modeling~\cite{yin2023continuous, chen2023crom}. INRs are known to be highly expressive and parameter-efficient, with the ability to represent high-resolution signals using very small neural networks~\cite{dupont2021coin, tancik2020fourier}, making them especially suitable for low-budget modeling tasks.

\subsection{Generalizing INRs Across Signals}
By design, INRs are optimized to represent a specific signal, such as a single image or scene, rather than to generalize across a dataset. As a result, a standard per-instance INR does not naturally generalize across a distribution of signals~\cite{VyaKus_Fit2025}. Several works address this limitation by introducing shared structures across INR instances. ~\cite{dupont2022functa} employed meta-learning~\cite{finn2017modelagnostic} to learn a shared INR capturing features common to the dataset, while each individual image is represented by image-specific activation modulations. Similarly,~\cite{vyas2025learningtrans} splits an INR into an encoder and a decoder component, where the shared encoder is trained jointly on a set of images, while each image-specific decoder is fine-tuned on an individual image.~\cite{wu2026disentangled} further separates shared factors from individual factors using a shared encoder-decoder pair with subject-specific encoders. These methods show that content sharing and meta-learning strategies can encourage INR representations to transfer across different signals. 

\subsection{Prediction based on INR representations}
Cross-signal consistency in INR representations motivates their use in downstream prediction tasks. Existing work can be divided into two groups depending on how the INR representation is utilized. 

The first group treats the INR itself as the representation of an input signal. Downstream predictions are performed on the INR's weights/learned weight embeddings~\cite{ICLR2024_c592fc7e, navon2023equivariant, zhou2023permutation, zhou2023neural} or function-level representations~\cite{dupont2022functa, bauer2023spatialfuncta}. However, although such signal-wise representations are well-suited for global tasks such as classification, they are less suitable for dense prediction, which requires predictions to be spatially localized.

The second group instead queries the INR at spatial coordinates and uses its coordinate-conditioned outputs or features for dense prediction. INRs are commonly used as prediction heads to convert image features generated by an external encoder into dense predictions, exploiting the continuous nature of INRs to make coordinate-conditioned predictions at arbitrary spatial locations.~\cite{yu2024q2aqueryingimp, you2023implicitan, 10.1007/978-3-031-72111-3_31, wei2024imedsam, sarkar2023parameter, 10.1007/978-3-031-16443-9_42}. Closer to our setting are methods in which the INR itself directly represents the signal, and dense predictions are derived directly from its queried outputs or hidden features~\cite{ranade2022ssdnerf, zarzar2022segnerf, zhi2021ilabel, Zhi_2021_ICCV, kohli2021semantic, kundu2022panoptic}. This has also been applied to various fields in medical imaging, including occupancy prediction~\cite{wysocki2026oscaroccu}, lesion segmentation~\cite{wu2026wsiinrimplicit}, and anatomical segmentation~\cite{VyaKus_Fit2025, stoltansnisf}. NISF~\cite{stoltansnisf} employs latent-conditioned INRs~\cite{park2019deepsdf} by using a shared INR across images and image-specific latent vectors. MetaSeg~\cite{VyaKus_Fit2025} further improves upon that by meta-learning a shared INR on a set of images that is adapted to each specific test image at test time. A segmentation head then takes in the hidden features of the adapted INR queried at different coordinates to generate the segmentation map. 

However, although these works have enabled INRs for dense prediction, a systematic characterization of how INR-based models compare with conventional architectures across parameter budgets, cross-site domain shifts, and augmentation settings is still lacking. In addition, how segmentation-relevant information is encoded across different INR layers remains largely unexplored.

\section{Methods}

\subsection{Preliminaries}

\textbf{Implicit Neural Representations}: an INR is a function $f_\theta(\cdot)$ parameterized by MLP parameters $\theta$ that is learned to map
$f_\theta(\cdot) : \mathbf{x} \mapsto I(\mathbf{x})$ for image $I \in \mathbb{R}^{H \times W}$ with height H and width W. \(\mathbf{x}\in \mathbb{R}^2\) is a coordinate in the image space and
\(I(\mathbf{x})\) is the pixel value at that coordinate. Let $h_{\theta}^{(l)}(\mathbf{x})$ denote the feature produced by the $l^{th}$ hidden layer of the INR $f_\theta(\cdot)$ queried at the input coordinate $\mathbf{x}$, with $l\in\{1,\ldots,L-1\}$, and $L$ the total number of INR layers.

\noindent\textbf{Meta-learned INR for MRI Segmentation:} MetaSeg~\cite{VyaKus_Fit2025} proposed a two-stage training framework that trains a SIREN INR~\cite{sitzmann2020implicit} $f_{\theta}(\cdot)$ with model parameters $\theta$ and an MLP segmentation head $g_{\phi}(\cdot)$ with parameters $\phi$. $f_{\theta}(\cdot):\mathbb{R}^2\rightarrow\mathbb{R}$ takes in coordinate $\mathbf{x}$ as input and is trained to reconstruct an image by predicting the corresponding intensity at coordinate $\mathbf{x}$. $g_{\phi}(\cdot)$ takes in the penultimate features $h_{\theta}^{(L-1)}(\mathbf{x})$ from the INR and is trained to predict the class label for coordinate $\mathbf{x}$. 

The first training stage is a MAML-style~\cite{finn2017modelagnostic} meta-learning stage that jointly trains the two components, with the INR $f_{\theta}(\cdot)$ being trained on reconstruction loss and the segmentation head $g_{\phi}(\cdot)$ on segmentation loss. For each training image, the inner loop adapts both $f_{\theta}(\cdot)$ and $g_{\phi}(\cdot)$ to reconstruct the image and predict its segmentation map, producing temporary image-specific parameters $\theta'$ and $\phi'$. The outer loop then updates the shared meta-learned initialization ${\theta^{m}, \phi^{m}}$ accordingly, where $m$ denotes the resulting meta-learned parameters from this stage.

In the second stage, the meta-learned INR $f_{\theta^{m}}(\cdot)$ is fit separately to each training image $I_i \in \{I_0,\ldots,I_n\}$, producing image-specific INR parameters $\theta^{I_i}$. After fitting, the adapted INR is queried at all image coordinates with its penultimate-layer features extracted, yielding the feature set $\{h_{\theta^{I_0}}^{(L-1)}(\mathbf{x}), \ldots, h_{\theta^{I_n}}^{(L-1)}(\mathbf{x})\}$. The meta-learned segmentation head $g_{\phi^{m}}(\cdot)$, using this feature set as input, is then further trained on the corresponding ground-truth segmentation maps. This produces the final segmentation head $g_{\phi^{*}}(\cdot)$, which is paired with the meta-trained INR $f_{\theta^{m}}(\cdot)$ from the first stage for inference.

At inference, given a test image $I$, the meta-trained INR $f_{\theta^{m}}(\cdot)$ is first fine-tuned to reconstruct $I$, producing image-specific parameters $\theta^{I}$. The adapted INR is then queried at each coordinate $\mathbf{x}$, and its penultimate-layer feature $h_{\theta^{I}}^{(L-1)}(\mathbf{x})$ is fed into the frozen segmentation head $g_{\phi^{*}}(\cdot)$ to predict its segmentation label.

\subsection{Motivating HierINRSeg: Analyzing Layer-wise Feature Geometries in MetaSeg}

To motivate our design, we start by investigating how semantics are encoded in a MetaSeg INR. Prior work suggests that INR frequency support expands with depth, while earlier layers emphasize lower-frequency components~\cite{yüce2022structured} and may be more transferable across signals~\cite{vyas2025learningtrans}. Here, we investigate whether a similar pattern holds for semantic information. Specifically, we ask the following question: do earlier INR layers also preserve transferable semantic structure, or is label-relevant information concentrated only in the penultimate features used by MetaSeg?

To answer this question, we performed two complementary analyses on the intermediate layers of a meta-trained MetaSeg INR. First, we examine how stable each layer is when fitting to a specific image. The intuition is that if a layer retains its global structure before and after fitting to one image, it may encode features that are less tied to that particular image and thus more transferable across images. In contrast, if a layer changes substantially after fitting, it is likely more specialized to the fitted image.

\begin{figure}[tb]
    \centering
    \includegraphics[width=0.9\linewidth]{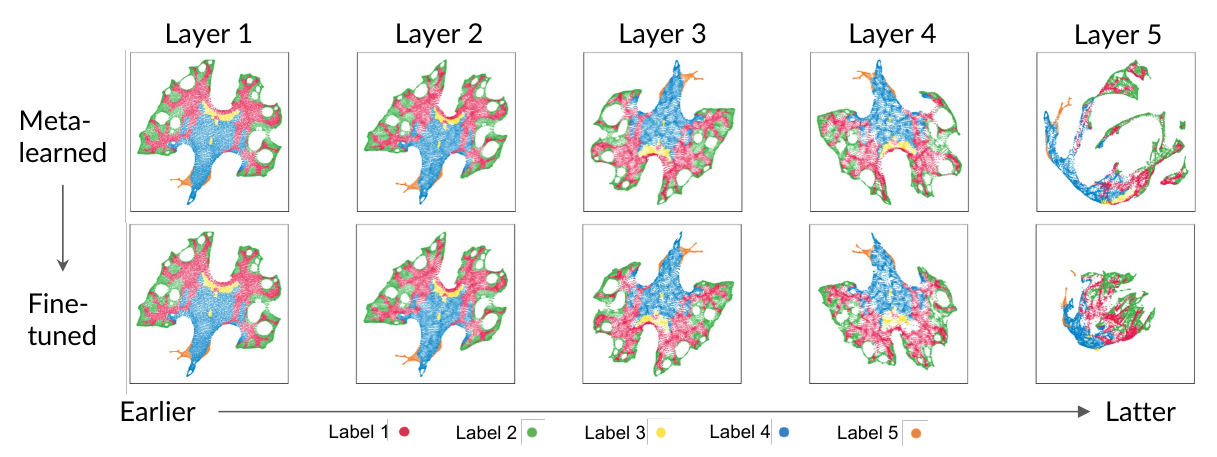}
    \caption{Visualizing MetaSeg INR features before and after fitting to a single image using UMAP. Earlier layers remain largely stable against fitting, while the final layer changes substantially, suggesting that early and middle layers contain more reusable structure while latter layers are more image-specific.}
    \label{fig:motiv0}
\end{figure}

For each hidden layer, we extract its features before and after fitting the INR to a given image. For the same layer, we jointly standardize the features before and after fitting, reduce them with Principal Component Analysis (PCA), retaining 90\% of the variance, and fit a shared UMAP~\cite{mcinnes2020umap} embedding. We then apply this embedding separately to the features before and after fitting. This allows us to compare the change in feature geometry for each layer in the same visualization space. As shown in Figure~\ref{fig:motiv0}, we observe that the final hidden layer changes substantially after fitting, while the earlier layers remain largely consistent. This suggests that earlier layers provide more stable representations, whereas the final hidden layer is more strongly affected by image-specific adaptation. Interestingly, we also observe that the earlier layers preserve the brain's spatial anatomy, which diminishes in the last layer. 

\begin{figure}[t]
    \centering
    \includegraphics[width=0.9\linewidth]{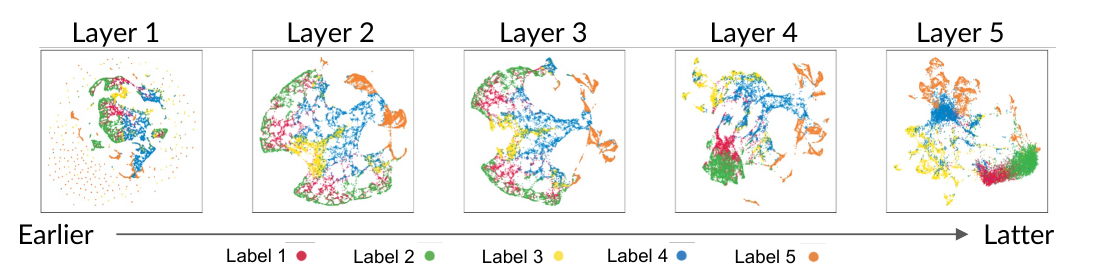}
    \caption{Layer-wise UMAP of MetaSeg INR features pooled across validation images after image-specific fitting. The first layer shows weak semantic organization, middle layers preserve a shared brain-like spatial structure, and latter layers become more label-compact while losing global anatomical layout.   }
    \label{fig:motiv1}
\end{figure}

Next, we further assess the features’ transferability across different images. We sample a class-balanced set of foreground pixels across the validation set and extract the same per-layer features after fine-tuning on each image. We then visualize each layer using the previous PCA and UMAP pipeline. As shown in Figure~\ref{fig:motiv1}, we observe a layer-wise transition in feature geometry. The first hidden layer shows neither clear label-wise grouping nor a coherent global anatomical structure, suggesting that it mainly represents low-level coordinate information not yet semantically organized across images. In contrast, the middle layers preserve a global structure resembling the spatial layout of the brain. Instead of forming isolated class clusters, the labels in these layers follow their anatomical allocation within the global structure. Since our images are spatially aligned, we interpret this as a shared spatial-anatomical organization. The latter layers gradually transition to the opposite: the global brain-like structure is no longer visible, but points from the same label become more locally compact. This suggests that latter features are more discriminative, while middle features retain richer anatomical topologies. To verify that these observations are not specific to a particular UMAP configuration, we repeat both analyses across multiple UMAP hyperparameter settings and observe consistent layer-wise trends, as shown in Appendix D and Figures~\ref{fig:umap_sweep_1} and~\ref{fig:umap_sweep_2}.

These observations suggest that segmentation-relevant information is distributed across the INR hierarchy. Earlier layers are more stable under image-specific fitting, middle layers preserve shared anatomy, and later layers provide more compact label-related features. Therefore, by using only the penultimate feature as in MetaSeg, we risk losing: (1) semantic information transferable across images. (2) spatial information about brain anatomy. This motivates our HierINRSeg, which fuses features from all intermediate INR layers with the segmentation head in a hierarchical fashion, allowing the segmentation head to combine complementary information from different INR levels.

\subsection{Hierarchical Feature Fusion}
Based on this analysis, we introduce HierINRSeg, which extends the MetaSeg architecture by leveraging multi-layer hidden features from the trained INR. Instead of relying only on the penultimate representation, HierINRSeg integrates INR features from different intermediate layers via hierarchical feature fusion, enabling the segmentation head to gradually combine low- to high-frequency information. 
 
\begin{figure}[bt]
    \centering
    \includegraphics[width=0.95\linewidth]{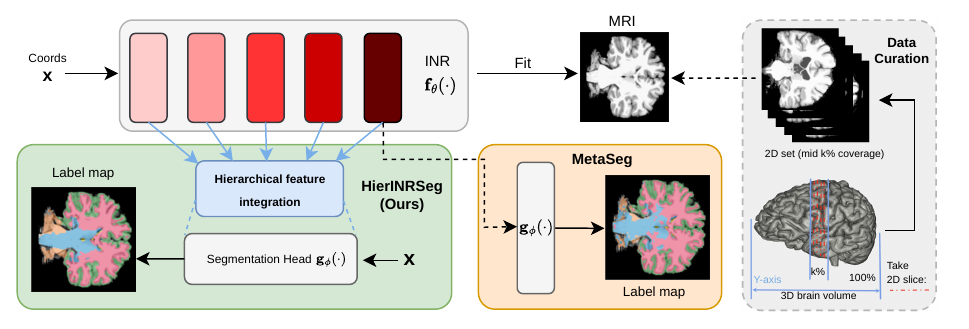}
    \caption{The HierINRSeg architecture: instead of predicting from a single INR layer as in MetaSeg (Orange panel), our HierINRSeg (green panel) progressively fuses multi-layer INR features with the segmentation head. This allows the segmentation head to gradually combine complementary spatial and semantic information from different INR layers before producing the final label map. The gray panel shows how 2D images are extracted from 3D MRI volumes. We select a segment along the Y-axis of the brain, centered at the brain center, that covers k\% of its total length. From this segment,
    slices in the XZ-plane are sampled to form the 2D images used in our experiments. We use k=5 for the main experiments, while the set of
    values for k is expanded to \{5, 20, 40, 60, 80\} in Section~\ref{sec:practical_scale}.}
    \label{fig:pipeline}
\end{figure}

Figure~\ref{fig:pipeline} provides an overview of the HierINRSeg architecture and highlights its differences from MetaSeg.

The SIREN INR $f_{\theta}^{}(\mathbf{\cdot})$ consists of five hidden sine layers of width $d$, followed by a
final linear layer for image intensity reconstruction~\cite{sitzmann2020implicit}. For each spatial
coordinate $\mathbf{x}\in\mathbb{R}^{2}$, it produces a scalar intensity
prediction together with intermediate hidden features
$h_{\theta}^{(l)}(\mathbf{x})$, where $l\in\{1,2,3,4,5\}$ and
$h_{\theta}^{(l)}(\mathbf{x})\in\mathbb{R}^{d}$. Here,
$h_{\theta}^{(l)}(\mathbf{x})$ denotes the feature vector at the $l$-th layer of the INR $f_{\theta}^{}(\mathbf{\cdot})$ parameterized by $\theta$.

The segmentation head $g_{\phi}(\cdot)$ predicts $c$ class logits for each input coordinate by progressively concatenating multi-level INR features with different layers of the segmentation head.
The first layer of $g_{\phi}(\cdot)$ is a SIREN layer that projects the coordinate $\mathbf{x}$ to a $d$-dimensional embedding $\mathbf{z}_0$. This follows the common practice of conditioning implicit segmentation decoders on coordinates/queries~\cite{wei2024imedsam, you2023implicitan, Reich2021, stoltansnisf}, while alternatives exist with no meaningful performance changes, as discussed in Appendix Section~\ref{appn:additional_anal}.
\[
\mathbf{z}_0 = \mathrm{embed}(\mathbf{x}),
\qquad \mathbf{z}_0 \in \mathbb{R}^{d}.
\]
With $[\cdot,\cdot]$ denoting concatenation, we define the fusion of two layer features
$\mathbf{a},\mathbf{b}\in\mathbb{R}^{d}$ as
\[
\operatorname{fuse}(\mathbf{a},\mathbf{b})
=
\mathrm{LN}\!\left(
\left[
\mathrm{LN}(\mathbf{a}),
\mathrm{LN}(\mathbf{b})
\right]
\right),
\qquad \operatorname{fuse}(\mathbf{a},\mathbf{b}) \in \mathbb{R}^{2d}
\]
Then, the hierarchical fusion process is
defined recursively for $l\in\{1,2,3,4\}$ as
\[
\mathbf{z}_{l}
=
\rho\!\left(
W_{l}\,\operatorname{fuse}\!\left(
\mathbf{z}_{l-1},
h_{\theta}^{(l)}(\mathbf{x})
\right)
+\mathbf{b}_{l}
\right),
\qquad \mathbf{z}_{l}\in\mathbb{R}^{d},
\]
where $\rho(\cdot)$ denotes a LeakyReLU activation, and $W_l$ and $\mathbf{b}_l$ denote
the learnable weight matrix and bias vector of the $l$-th fusion layer,
respectively. Each intermediate fusion layer therefore maps
$\mathbb{R}^{2d}\rightarrow\mathbb{R}^{d}$.

Finally, the segmentation logits for $c$ classes are produced by fusing the last segmentation
state with the deepest INR feature:
\[
g_{\phi}\!\left(
\mathbf{x}
\right)
=
W_{\mathrm{out}}\,
\operatorname{fuse}\!\left(
\mathbf{z}_{4},
h_{\theta}^{(5)}(\mathbf{x})
\right)
+
\mathbf{b}_{\mathrm{out}}
\in
\mathbb{R}^{c},
\]
where $W_{\mathrm{out}}$ and $\mathbf{b}_{\mathrm{out}}$ are the learnable weight matrix and bias vector of the output layer.

\section{Experiments}

\subsection{Datasets and Evaluation}
\label{sec:D&E}
We used the MRI scans from the alphabetically first three centers in the public ABIDE I~\cite{di2014autism} dataset: Caltech, CMU, and KKI. Each 3D volume is RAS-aligned, cropped to the nonzero brain bounding box with a small padding, and MRI intensities are normalized to 0-1 using the 1st and 99th percentile values within the brain region. For the main experiments, we select the central 5\% of slices along the Y-axis and resample them to $160^2$ (an illustration of the slicing process is shown in Figure~\ref{fig:pipeline}). The segmentation label maps are merged into one background/ignored class and five foreground classes.   

We evaluate in both in-domain (ID) and out-of-domain (OOD) settings. KKI, the largest set (491 slices), is used as the source dataset and split at the subject level into 70\%/15\%/15\% train/validation/test splits. All slices from a given subject were kept in the same split, ensuring no subject overlap across splits. Caltech (386 slices) and CMU (283 slices) are used only as OOD test datasets. Using two target centers produces two independent OOD evaluations, reducing the chance that the conclusion is driven by specific characteristics of a single target site. Training is always performed on the KKI training split, ID performance is evaluated on the KKI test split, and OOD performance is evaluated on all images from Caltech and CMU. We report the mean Dice averaged over subjects and foreground classes.

\subsection{Implementation Details}

In this work, we primarily compare our method against MetaSeg~\cite{VyaKus_Fit2025} and U-Nets~\cite{ronneberger2015unet} across training settings and parameter budgets. In addition, we compare with the state-of-the-art convolution-based nnUNet-v2~\cite{isensee2021nnu, isensee2024nnunet}, the transformer-based TransUnet~\cite{chen2021transu}, and UNeXT~\cite{valanarasu2022unext}, a highly parameter-efficient U-Net variant. All runs are repeated using three random seeds, with the mean and standard deviation of Dice scores reported.

For U-Net baselines, we use a standard 2D five-level architecture, with channel widths varied to match different parameter budgets. The training for MetaSeg and HierINRSeg follows the two-stage MetaSeg procedure. The number of outer-loop iterations in the first stage is capped at 5000, with the INR/segmentation head selected from the iteration with the highest validation Dice. For the second step, the segmentation head is trained for up to 4000 epochs, with early stopping after 5 consecutive non-improving validation rounds. Detailed hyperparameters are provided in Appendix Section~\ref{sec:impl_details}. 

\begin{figure*}[t]
    \centering

    \includegraphics[width=0.75\textwidth]{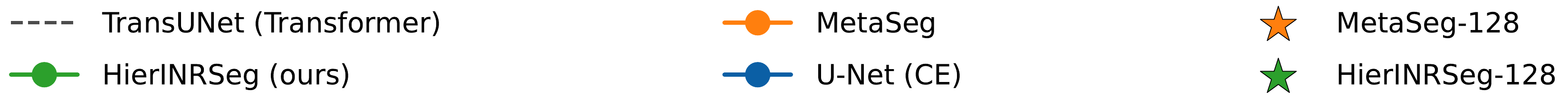}

    \noindent
    \begin{minipage}[t]{0.327\textwidth}
        \centering
        \includegraphics[width=\linewidth]{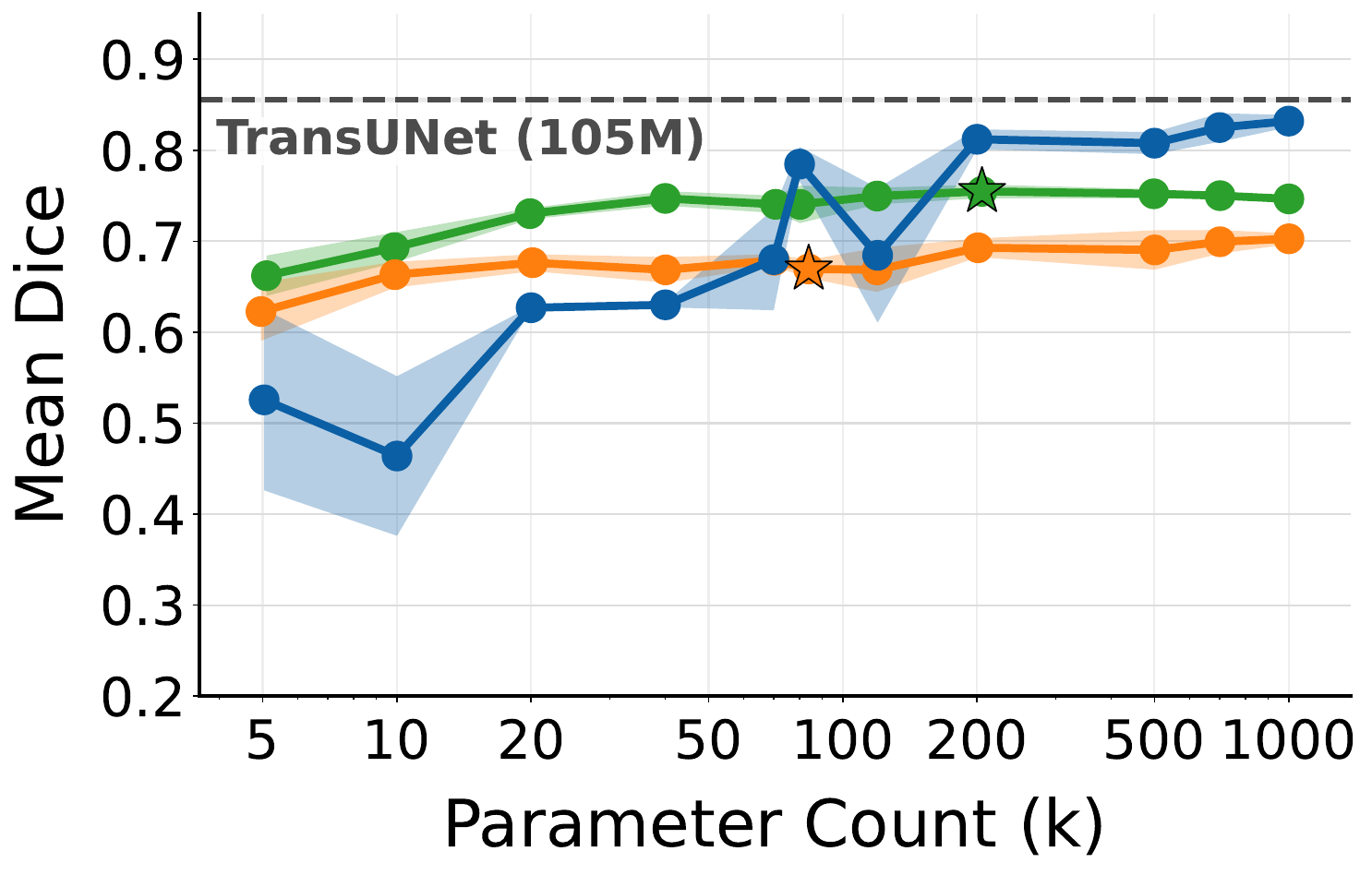}
        {Testing: (a) KKI (ID)}
    \end{minipage}%
    \begin{minipage}[t]{0.31\textwidth}
        \centering
        \includegraphics[width=\linewidth]{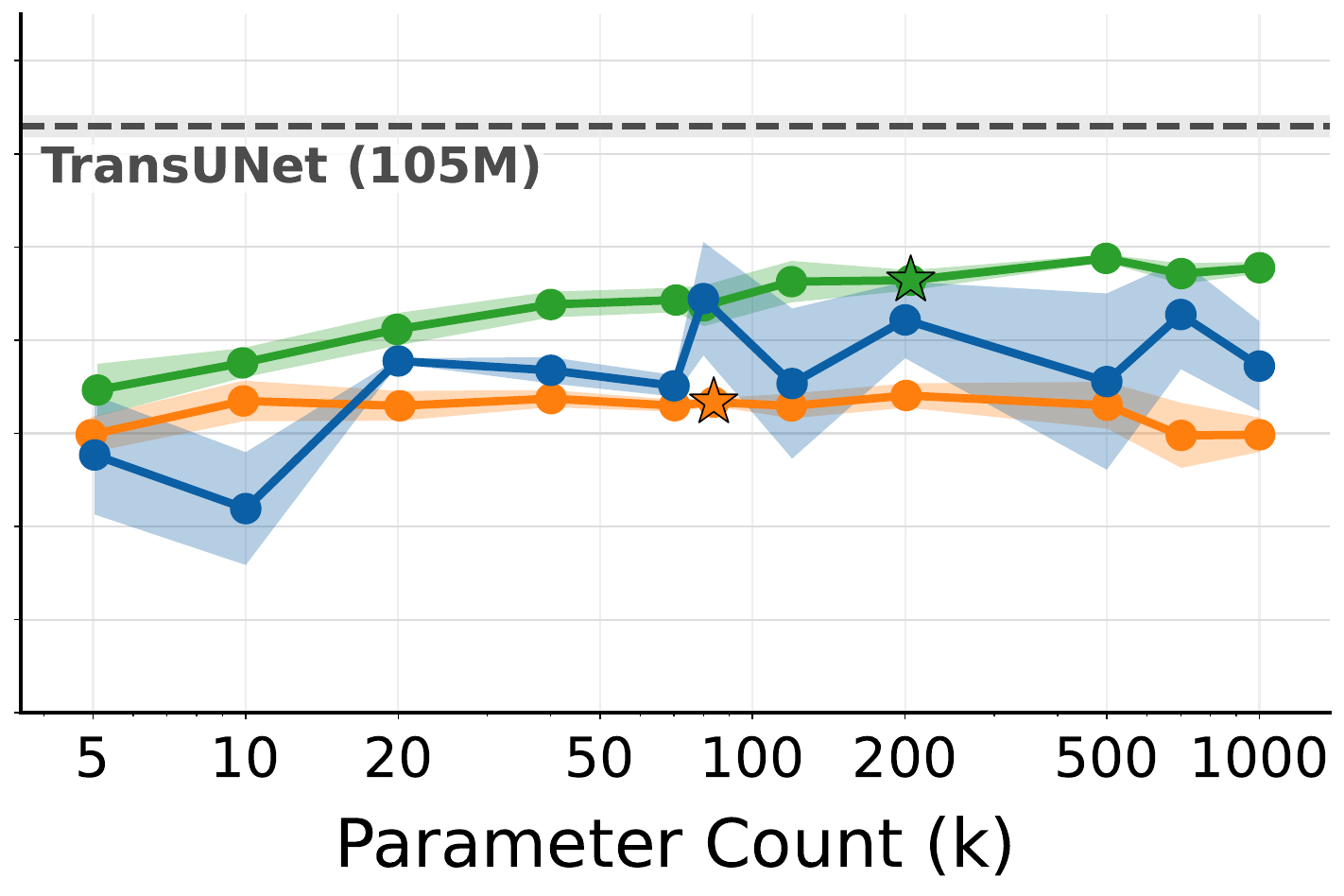}
        {(b) Caltech (OOD)}
    \end{minipage}%
    \begin{minipage}[t]{0.31\textwidth}
        \centering
        \includegraphics[width=\linewidth]{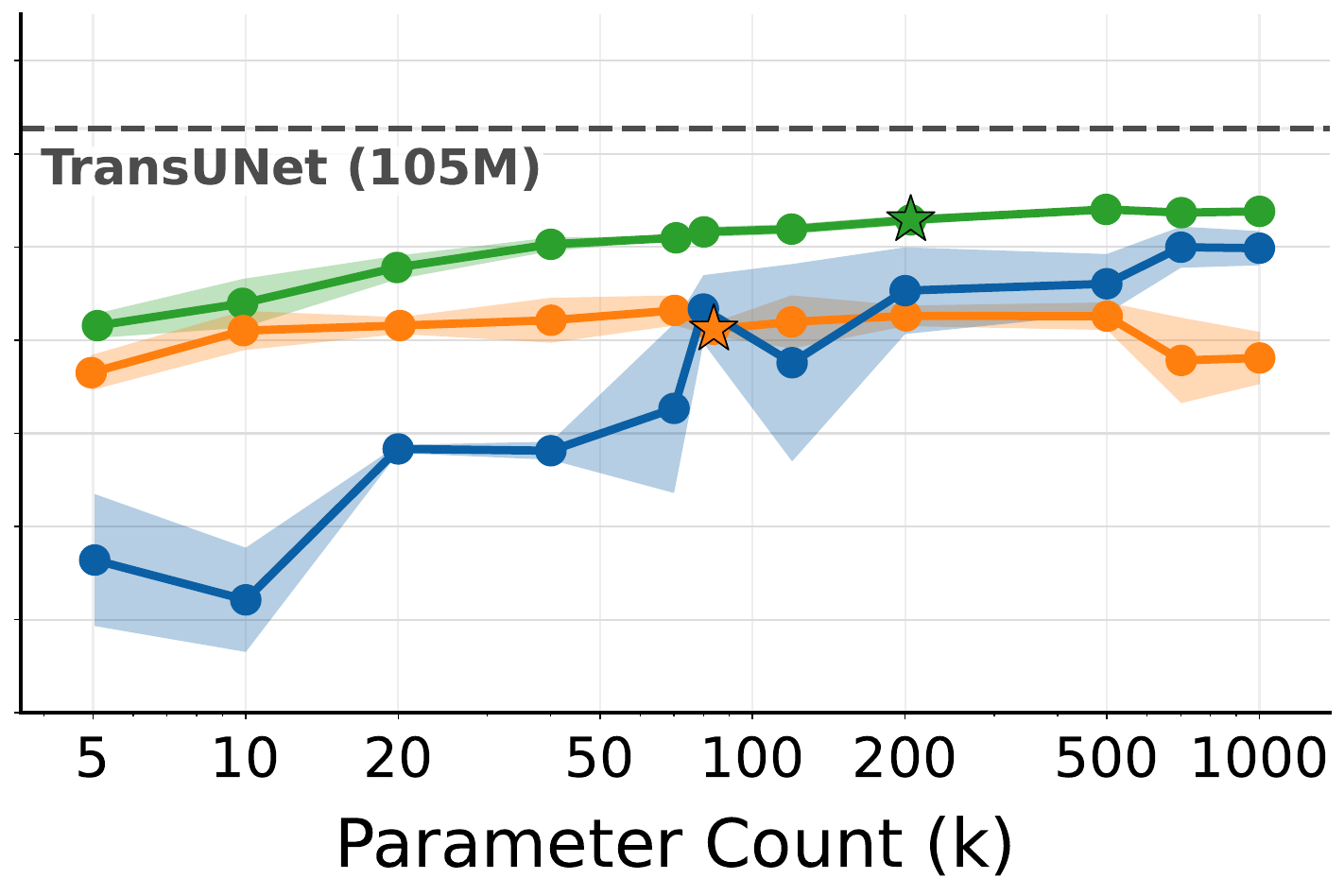}
        {(c) CMU (OOD)}
    \end{minipage}

    \vspace{0.5em}

    \caption{Controlled parameter-scaling comparison. All models are trained on KKI without augmentation and evaluated on KKI (ID, test split) and Caltech / CMU (OOD).  We compare MetaSeg, HierINRSeg, and U-Nets across parameter budgets from 5K to 1M.  Dashed lines represent the large-scale TransUNet. All runs are repeated for three different seeds, with markers denoting the mean and the surrounding shades showing ± standard deviation. The two stars represent MetaSeg and our HierINRSeg using MetaSeg's original hidden dimension of 128. HierINRSeg consistently improves over MetaSeg across parameter budgets, while U-Net requires substantially larger capacity to become competitive, especially on the OOD test sets.}
    \label{fig:main_noaug}
\end{figure*}

\subsection{Comparison Protocols}
\label{sec:protocols}
We report two comparison protocols. For the \textbf{controlled scaling comparison}, all models are directly trained under the same basic setting. We evaluate each architecture across 11 parameter budgets ranging from 5K to 1M parameters. For each budget, we construct corresponding versions of U-Net, MetaSeg, and HierINRSeg. For U-Net, we vary the number of feature channels at each resolution level to control the model size. For the INR-based models, we vary the hidden-layer dimensions of the INR and segmentation head. This protocol provides a controlled comparison of how each architecture scales with model capacity. 

For the \textbf{practical comparison} protocol, as U-Nets are known to benefit substantially from augmentation, for a stronger comparison, we repeat the experiments with additional augmentation applied to the training set across all models. Each image undergoes a randomly composed set of standard intensity augmentations, including gamma transformation, contrast scaling, intensity shift, and Gaussian noise. Augmentation details are listed in Appendix Section~\ref{sec:impl_details}. In addition, we train a second set of U-Net baselines using the same focal loss as MetaSeg and HierINRSeg, allowing us to separate architectural differences from the effect of the loss function. This protocol provides a more rigorous comparison in which the U-Net baselines are given commonly used robustness improvements. Furthermore, a UNeXt parameter sweep is provided under the same augmentation scheme and parameter budgets to compare INR-based models with U-Net-based architectures optimized for parameter efficiency.

\section{Results and Analysis}

\subsection{Controlled scaling comparison}
Figure~\ref{fig:main_noaug} reports the results under the \textbf{controlled scaling comparison} setting. HierINRSeg consistently outperforms the MetaSeg baseline at the same parameter budget in both in-domain and out-of-domain evaluations. Compared with U-Net, the relative performance depends more strongly on model size. On the in-domain KKI test set, HierINRSeg consistently outperforms U-Net at parameter counts up to 50K, while U-Net surpasses HierINRSeg once the parameter budget reaches 200K. In the out-of-domain setting, HierINRSeg performs better than U-Net across all parameter budgets on CMU, and across almost all budgets on Caltech, with the only exception at 80K, where it is slightly below U-Net.

The INR-based methods show a very different scaling pattern from U-Nets. Both MetaSeg and HierINRSeg start from a much stronger performance level in the low-parameter regime and remain stable as the parameter budget increases, with only moderate improvements. In contrast, U-Net starts from substantially lower performance at small parameter budgets but benefits more from increasing model capacity. U-Net also shows much larger variation between seeds, especially in the low-budget regime. These results suggest that meta-trained INR representations provide a strong inductive bias for compact segmentation models, whereas U-Nets require larger capacity to become competitive.

\begin{figure*}[t]
    \centering

    \includegraphics[width=0.75\textwidth]{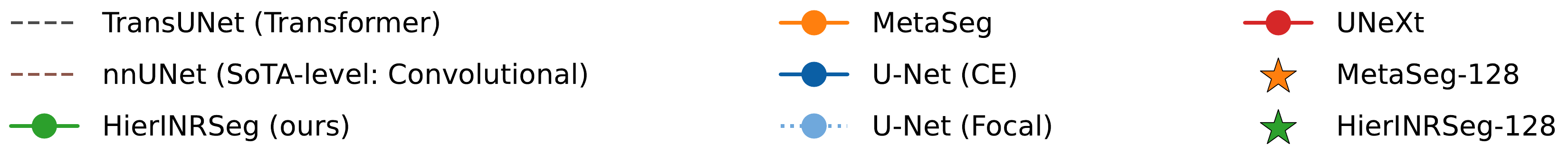}

    \noindent
    \begin{minipage}[t]{0.327\textwidth}
        \centering
        \includegraphics[width=\linewidth]{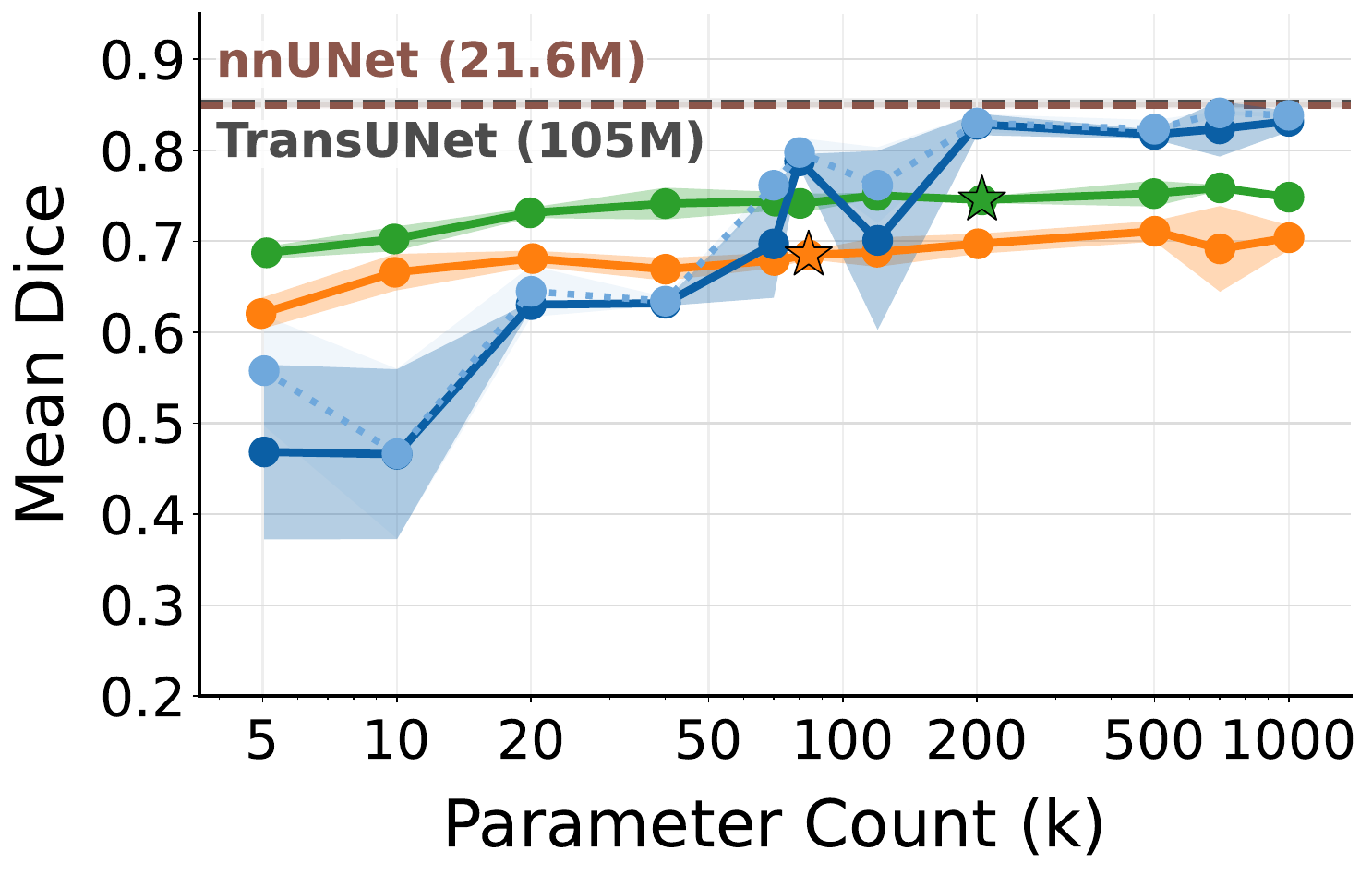}
    \end{minipage}%
    \begin{minipage}[t]{0.31\textwidth}
        \centering
        \includegraphics[width=\linewidth]{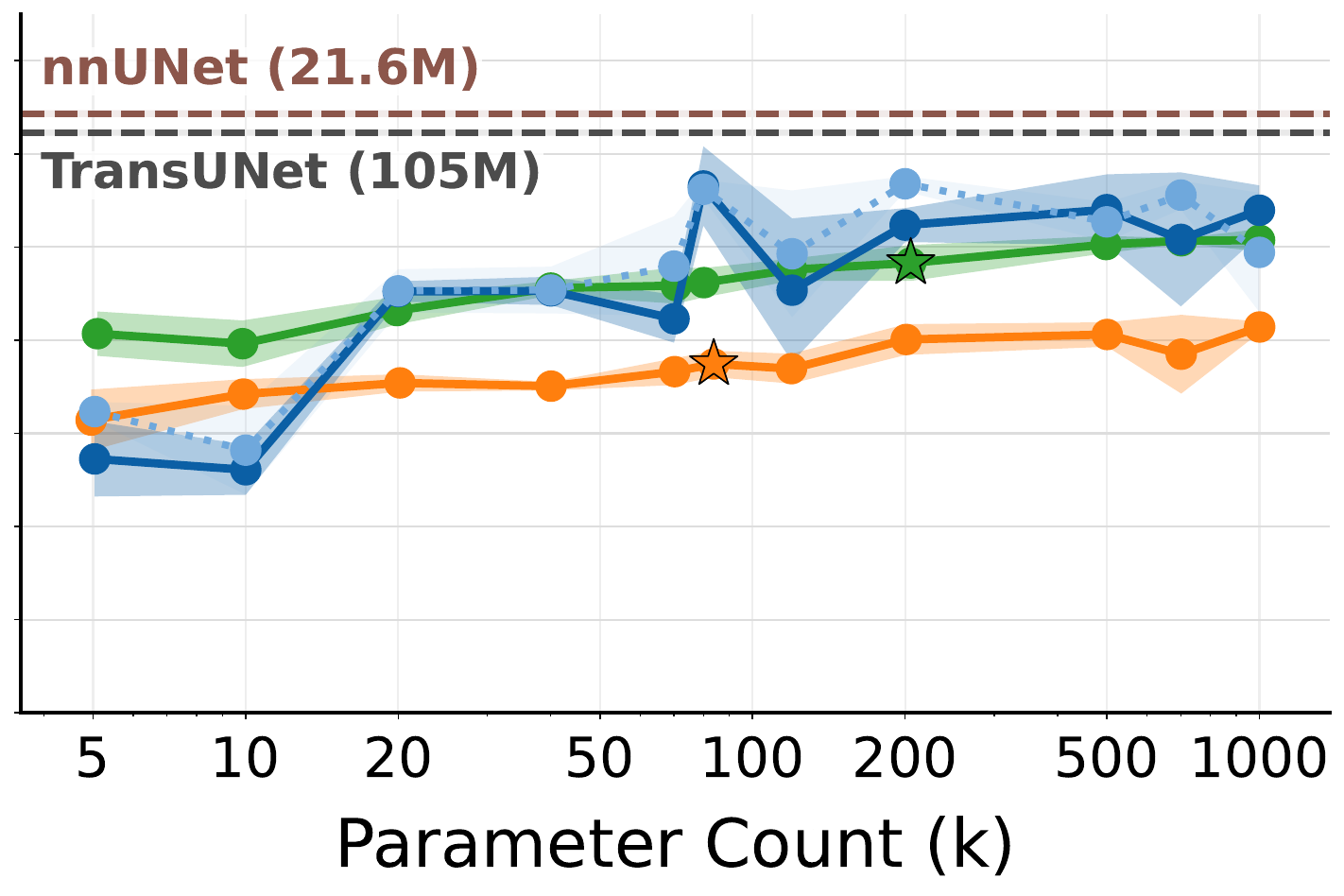}
    \end{minipage}%
    \begin{minipage}[t]{0.31\textwidth}
        \centering
        \includegraphics[width=\linewidth]{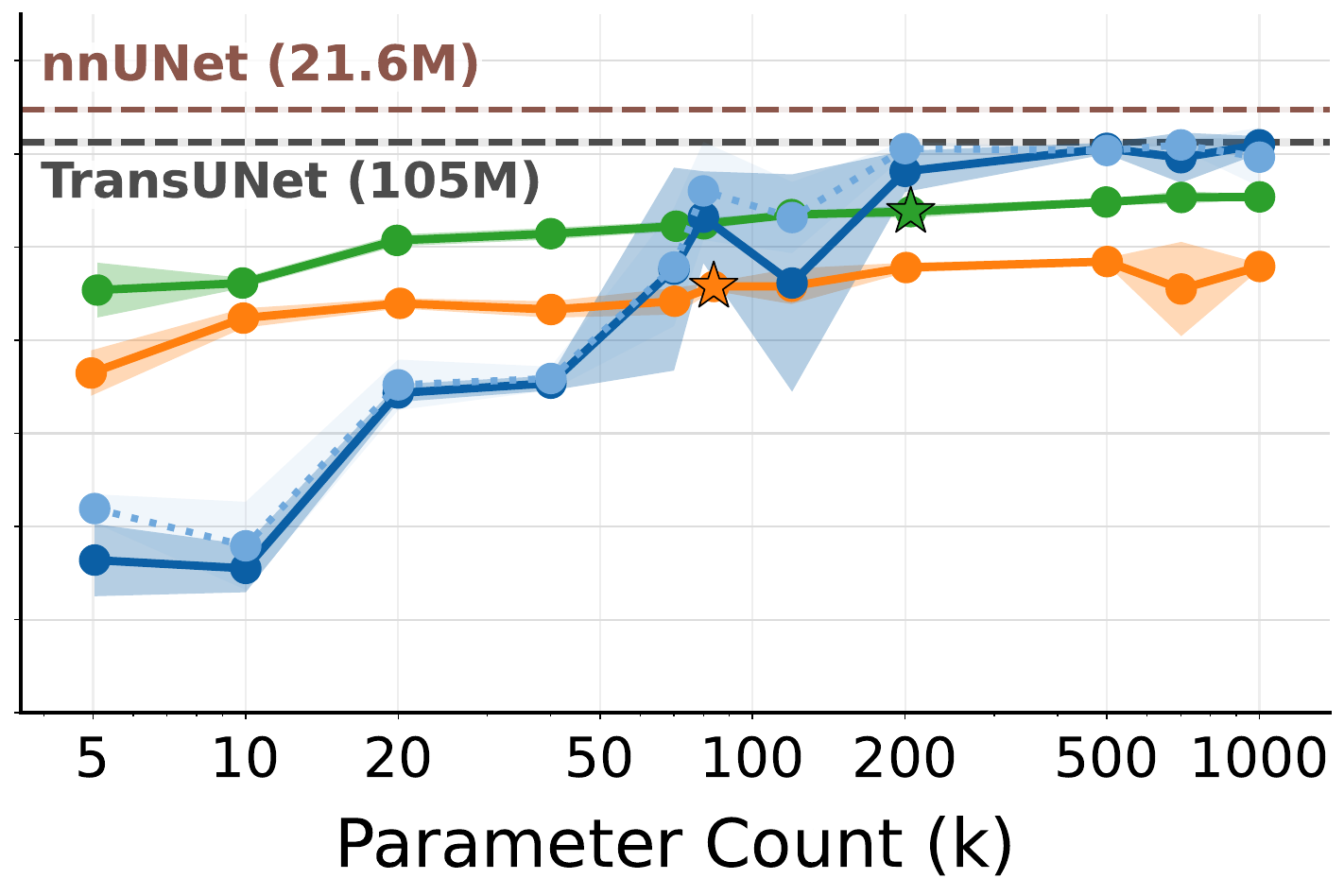}
    \end{minipage}

    \begin{minipage}[t]{0.327\textwidth}
        \centering
        \includegraphics[width=\linewidth]{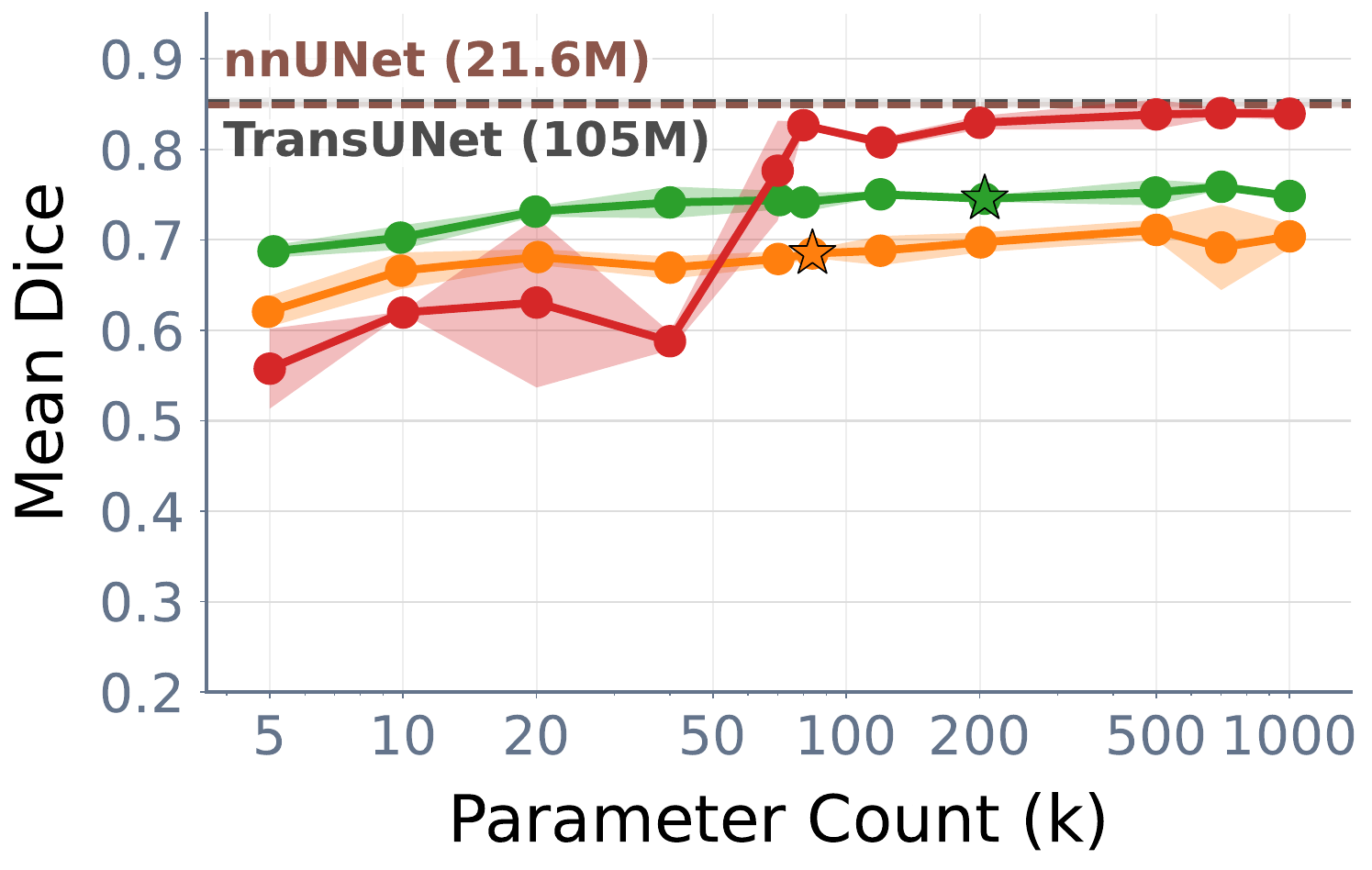}
        {Testing: (a) KKI (ID)}
    \end{minipage}%
    \begin{minipage}[t]{0.31\textwidth}
        \centering
        \includegraphics[width=\linewidth]{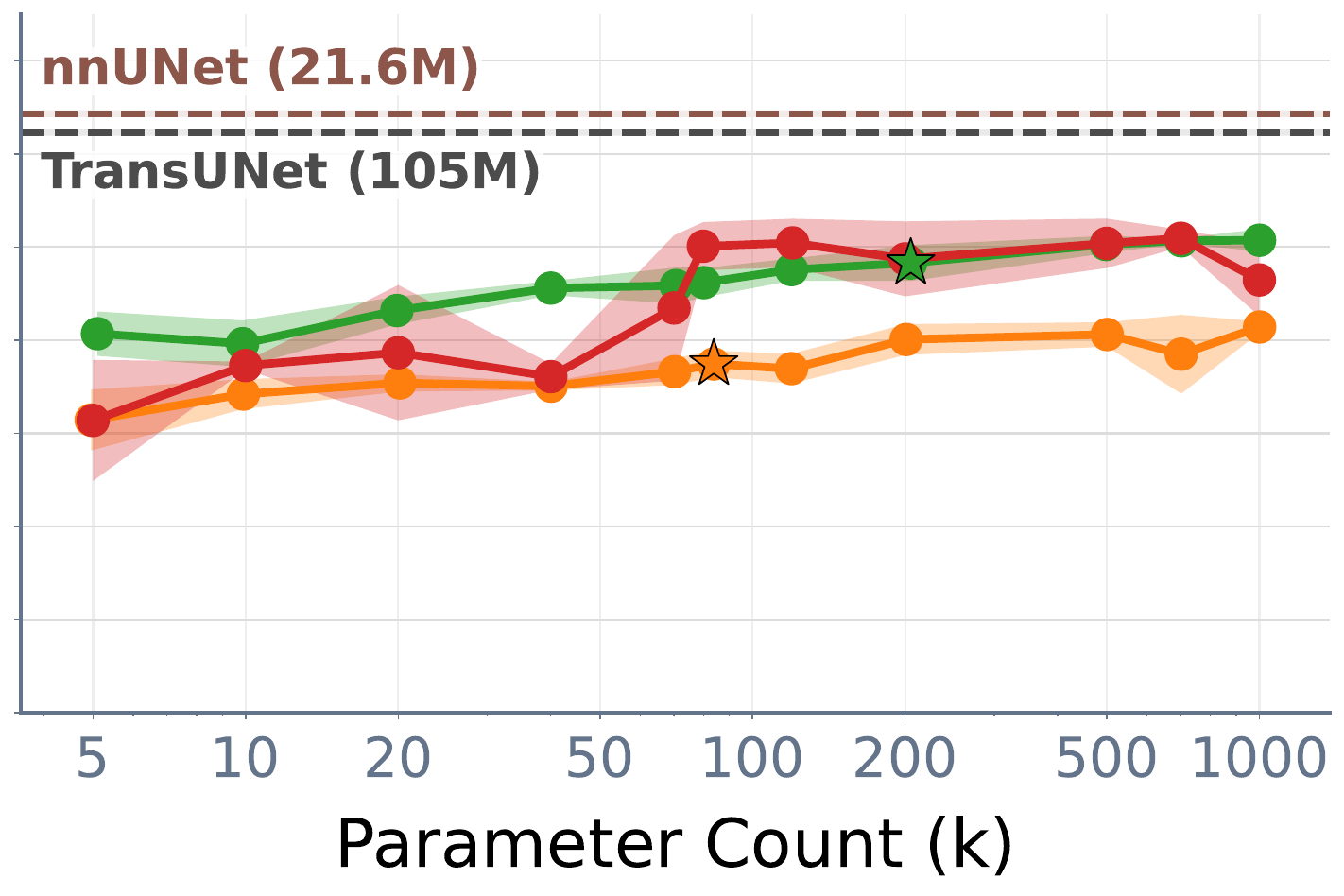}
        {(b) Caltech (OOD)}
    \end{minipage}%
    \begin{minipage}[t]{0.31\textwidth}
        \centering
        \includegraphics[width=\linewidth]{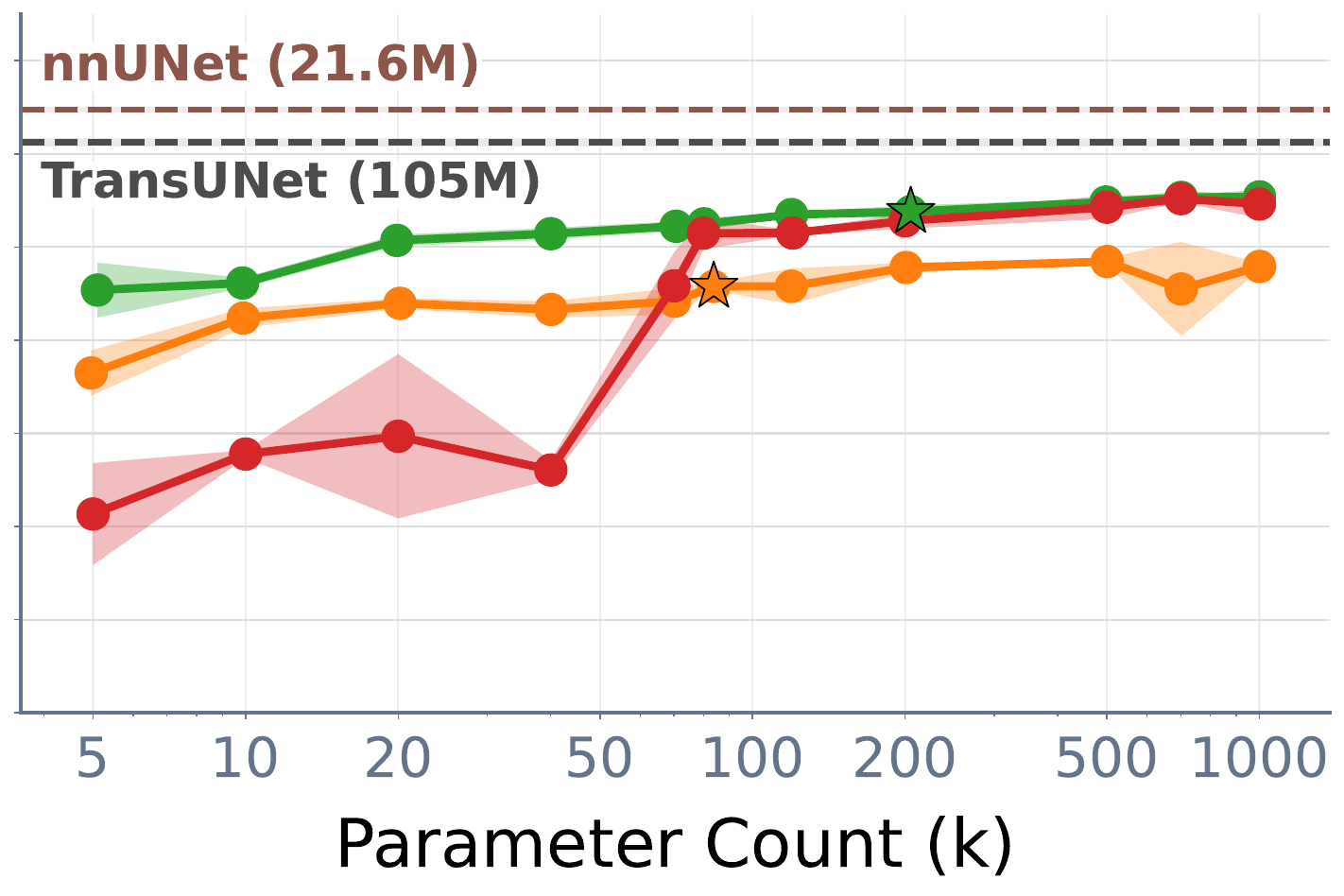}
        {(c) CMU (OOD)}
    \end{minipage}

    \vspace{0.5em}

    \caption{Practical parameter-scaling comparison with strengthened training. All models are trained on KKI with augmentation and evaluated on KKI (ID, test split) and Caltech/CMU (OOD). We include both cross-entropy and focal-loss U-Nets and the parameter-efficient UNeXt, with the large-scale nnU-Net and TransUNet shown as reference baselines. All runs are repeated for three different seeds, with markers denoting the mean and the surrounding shades showing ± standard deviation. Our HierINRSeg consistently improves over MetaSeg and remains competitive with strengthened U-Net and UNeXt baselines across domains. The UNeXt parameter sweep is shown in the second row for visual clarity.}
    \label{fig:main_aug}
\end{figure*}

\subsection{Practical scaling comparison with strengthened training}
\label{sec:practical_scale}

Figure~\ref{fig:main_aug} reports the results under the \textbf{practical comparison} setting. In the in-domain setting, the overall trends remain similar to the controlled comparison. The performance of all INR models changes only modestly, and HierINRSeg continues to outperform MetaSeg across all parameter budgets. The relative comparison with U-Net also remains largely unchanged: INR-based models are stronger in the low-parameter regime, while U-Net becomes more competitive as model capacity increases. Training U-Net with focal loss provides a small improvement over cross-entropy, but does not substantially change the overall ranking. This trend is also observed for UNeXt, which outperforms standard U-Nets at the lowest parameter budgets and remains well below HierINRSeg across the lower-budget regime, but surpasses HierINRSeg at larger budgets. 

The effect of the practical training protocol is more notable in the out-of-domain setting. Although augmentation has only a limited effect on the INR-based methods, it substantially improves U-Net performance on both target datasets. As a result, on the Caltech OOD test set, strengthened U-Nets achieve a performance level similar to HierINRSeg at medium and large parameter budgets, while HierINRSeg retains a clear advantage at very small budgets. On CMU, the comparison becomes similar to the in-domain setting: all U-Nets remain below HierINRSeg at budgets below 80K, but surpass it once the parameter count reaches approximately 200K. Focal loss gives an additional but smaller improvement over cross-entropy, suggesting that most of the OOD gain comes from augmentation. UNeXt exhibits relatively weaker generalization, with HierINRSeg outperforming UNeXt at lower parameter counts, while the two methods achieve similar performance at higher budgets.

\begin{figure}[tb]
    \centering
    \includegraphics[width=0.7\linewidth]{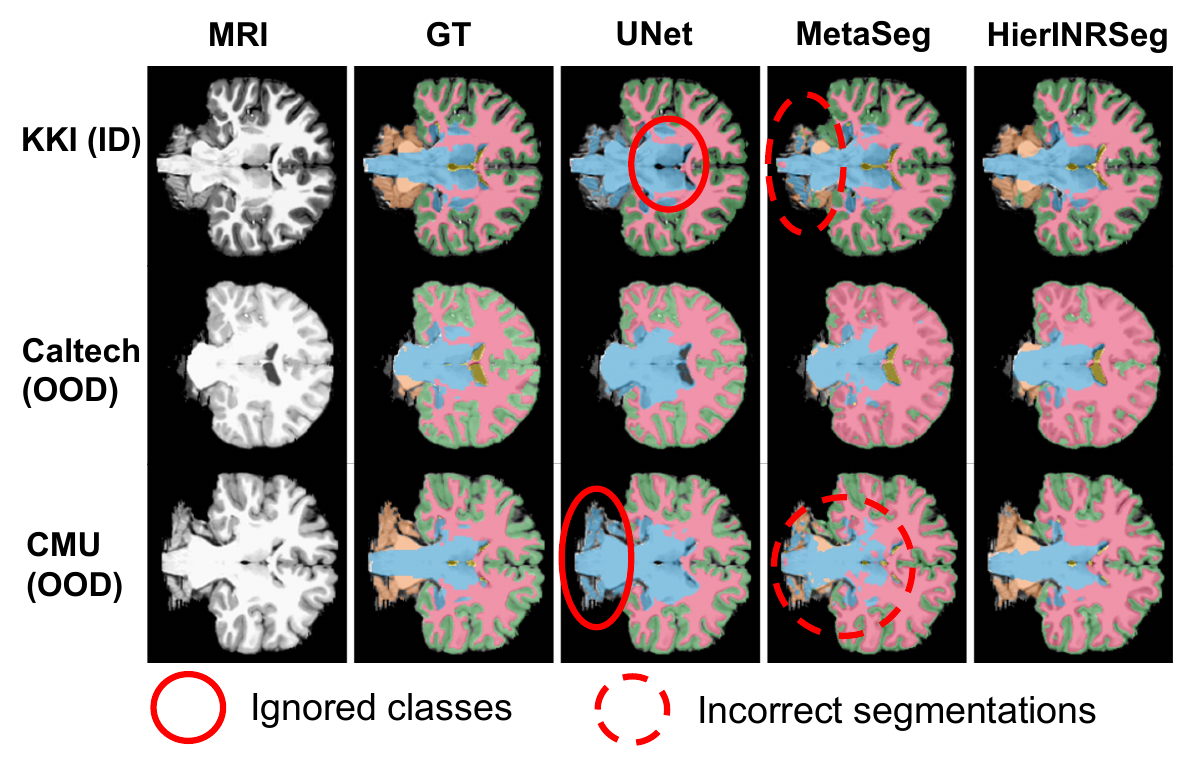}
    \caption{Qualitative results at the 20K parameter budget with training augmentation. We compare predictions from U-Net, MetaSeg, and HierINRSeg on the in-domain KKI test set and the out-of-domain Caltech and CMU test sets. Solid red circles indicate ignored regions, and dashed red circles indicate incorrect segmentations. HierINRSeg produces segmentations that more closely match the ground truth across both in-domain and out-of-domain examples.}
    \label{fig:res_exp_20k}
\end{figure}

These results refine the conclusion from the controlled comparison. Our HierINRSeg remains consistently stronger than MetaSeg and is especially effective in the low-budget regime. However, U-Nets benefit more from standard augmentation, narrowing or reversing the gap at larger parameter budgets. This suggests that meta-trained INR models do not benefit as much from the same augmentation strategy as U-Net-based baselines. Figure~\ref{fig:res_exp_20k} provides qualitative examples of the segmentation performance of each model at a 20k parameter budget for all three test sets. We observe that HierINRSeg produces segmentations that more closely match the ground truth while reducing common errors seen in other models, such as missed anatomical structures and mislabeled tissues.

\textbf{Sensitivity to 3D spatial coverage}. Beyond domain shift and parameter/augmentation budgets, we examine robustness to increasing spatial coverage within each 3D brain volume. In addition to the central 5\% of slices used in the main experiments, we evaluate HierINRSeg, MetaSeg, and U-Net on KKI by sampling slices from progressively wider central regions spanning 20\%, 40\%, 60\%, and 80\% of each 3D brain volume (see \textit{data curation} panel in Figure~\ref{fig:pipeline} for an illustration), while preserving the subject-level splits. Broader slice coverage includes anatomically more diverse locations, therefore introducing greater within-dataset spatial variability. We evaluate two representative budgets, 10K and 200K, corresponding to relatively low- and higher-capacity regimes identified in our parameter-scaling experiments. As shown in Table~\ref{tab:spatial_variability}, at the higher 200K budget, U-Net retains its advantage and slightly improves as sampling range increases, while both INR-based models show some performance degradation. In contrast, at the lower 10K budget, HierINRSeg consistently outperforms U-Net across all slice ranges, with the gap becoming noticeably smaller around ranges of 60\%-80\%. Importantly, HierINRSeg outperforms MetaSeg across both budgets and all slice ranges, while showing substantially lower sensitivity to increasing spatial coverage. This further highlights the advantages of utilizing multi-layer INR features, but also reveals the limitations of INR-based models when handling broader spatial coverage and the associated increase in spatial variability. The out-of-domain datasets show largely consistent trends, with the main difference being that U-Net slightly surpasses HierINRSeg at 80\% spatial coverage under the 10K budget, as shown in Appendix Tables~\ref{tab:spatial_variability_caltech} and ~\ref{tab:spatial_variability_cmu}.

\begin{table}[b]
\centering
\caption{KKI in-domain performance under increasing 3D spatial coverage. Slices are sampled from progressively wider central regions of each volume (5--80\%). Results are reported at representative 10K and 200K parameter budgets.}
\label{tab:spatial_variability}
\small
\setlength{\tabcolsep}{5pt}
\begin{tabular}{llccccc}
\toprule
\multirow{2}{*}{Budget} & \multirow{2}{*}{Model} 
& \multicolumn{5}{c}{Performance (Dice) on each percentage} \\
\cmidrule(lr){3-7}
& & 5\% & 20\% & 40\% & 60\% & 80\% \\
\midrule
\multirow{3}{*}{10K}
& MetaSeg    & 0.666 ± 0.020 & 0.648 ± 0.016 & 0.616 ± 0.011 & 0.535 ± 0.034 & 0.526 ± 0.016\\
& HierINRSeg &\textbf{ 0.703} ± 0.013 & \textbf{0.724} ± 0.059 & \textbf{0.680 }± 0.008 & \textbf{0.615} ± 0.019 & \textbf{0.644 }± 0.024 \\
& U-Net      & 0.466 ± 0.093 & 0.450 ± 0.061 & 0.451 ± 0.016 & 0.476 ± 0.090 & 0.617 ± 0.014\\
\midrule
\multirow{3}{*}{200K}
& MetaSeg    & 0.697 ± 0.011 & 0.694 ± 0.018 & 0.623 ± 0.007 & 0.512 ± 0.055 & 0.517 ± 0.015 \\
& HierINRSeg & 0.746 ± 0.003 & 0.768 ± 0.005 & 0.735 ± 0.048 & 0.666 ± 0.016 & 0.640 ± 0.034 \\
& U-Net      & \textbf{0.829} ± 0.012 & \textbf{0.857 }± 0.008 & \textbf{0.855} ± 0.010 & \textbf{0.844} ± 0.010 & \textbf{0.869} ± 0.013 \\
\bottomrule
\end{tabular}
\end{table}

\subsection{Analyzing INRs' Sensitivity to Augmentation}
\label{sec:aug_sens_anal}

To understand why INR-based models benefit only modestly from train-time augmentation, we studied the inference-time fitting process of the INRs. This adaptation step is unique to INRs and thus a likely source of their different behavior from U-Nets. For each of 30 randomly selected KKI validation images $I$, we generate an augmented version $A(I)$ using the same augmentation scheme as our \textit{practical scaling} experiments (see Section~\ref{sec:protocols}). Then, we select a different image $I'$ to serve as a reference. Starting from the same meta-trained (MetaSeg/HierINRSeg) checkpoint, we separately adapt each INR to $I$, $A(I)$, and $I'$, and measure the relative augmentation effect at layer $l$ after the $k^{th}$ fitting step as
\begin{equation}
R_l(k) = \frac{1 - S_l(I_k, A(I)_k)}{1 - S_l(I_k, I'_k)},
\label{eq:r}
\end{equation}
where $S_l(\cdot,\cdot)$ is the mean cosine similarity between hidden features at layer $l$. The denominator acts as a normalization term, measuring augmentation sensitivity relative to the effect of switching to a different image. A small $R_l(k)$ indicates that augmentation changes the representation much less than replacing the image with another sample. In contrast, a large $R_l(k)$ means that the augmentation-induced change is comparable to the inherent variations between different images.

\begin{figure}[t]
    \centering
    \noindent
    \begin{minipage}[t]{0.38\linewidth}
        \centering
        \includegraphics[width=\linewidth]{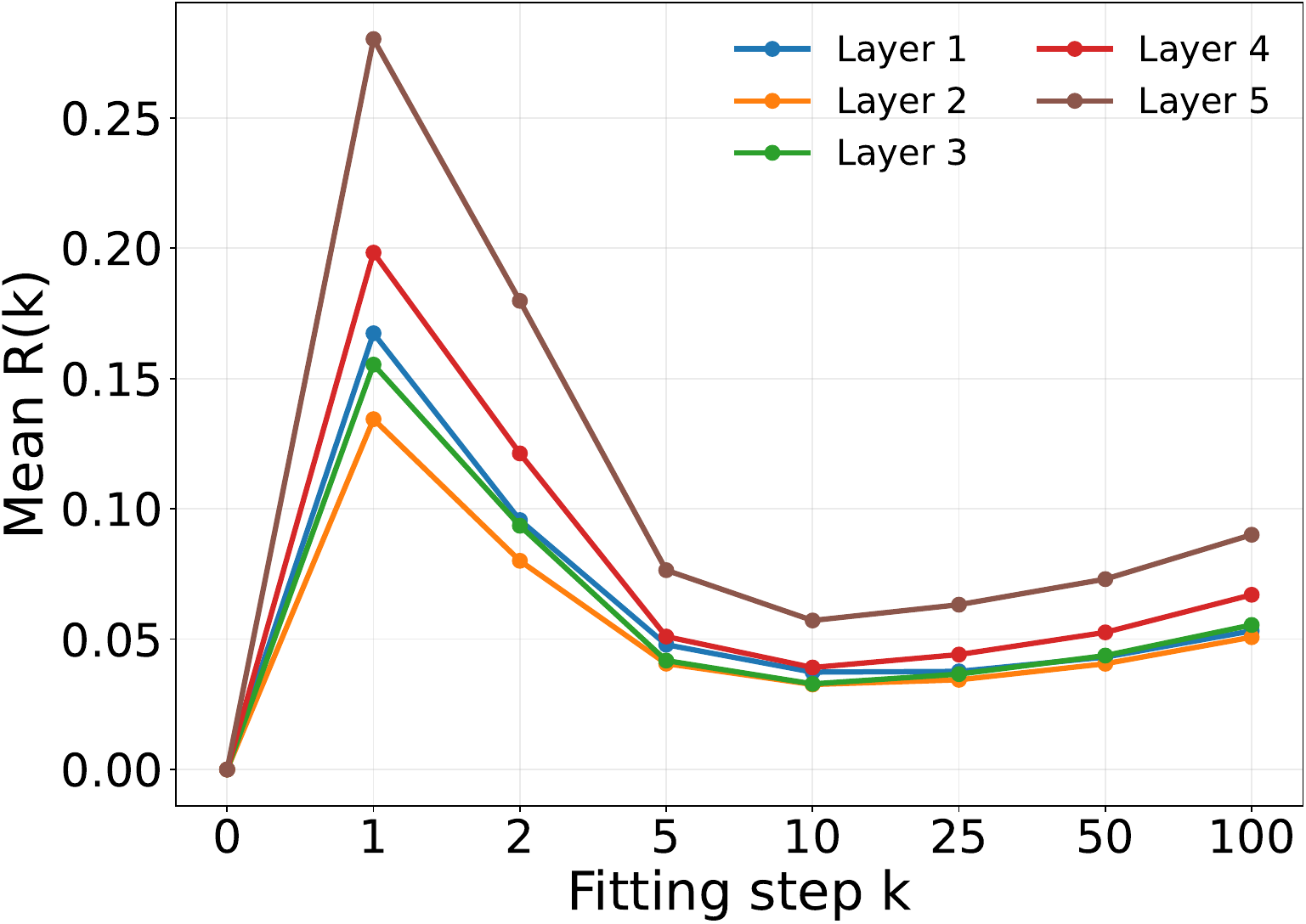}
        {Model: (a) HierINRSeg}
    \end{minipage} 
    \hspace{2em}
    \begin{minipage}[t]{0.38\linewidth}
        \centering
        \includegraphics[width=\linewidth]{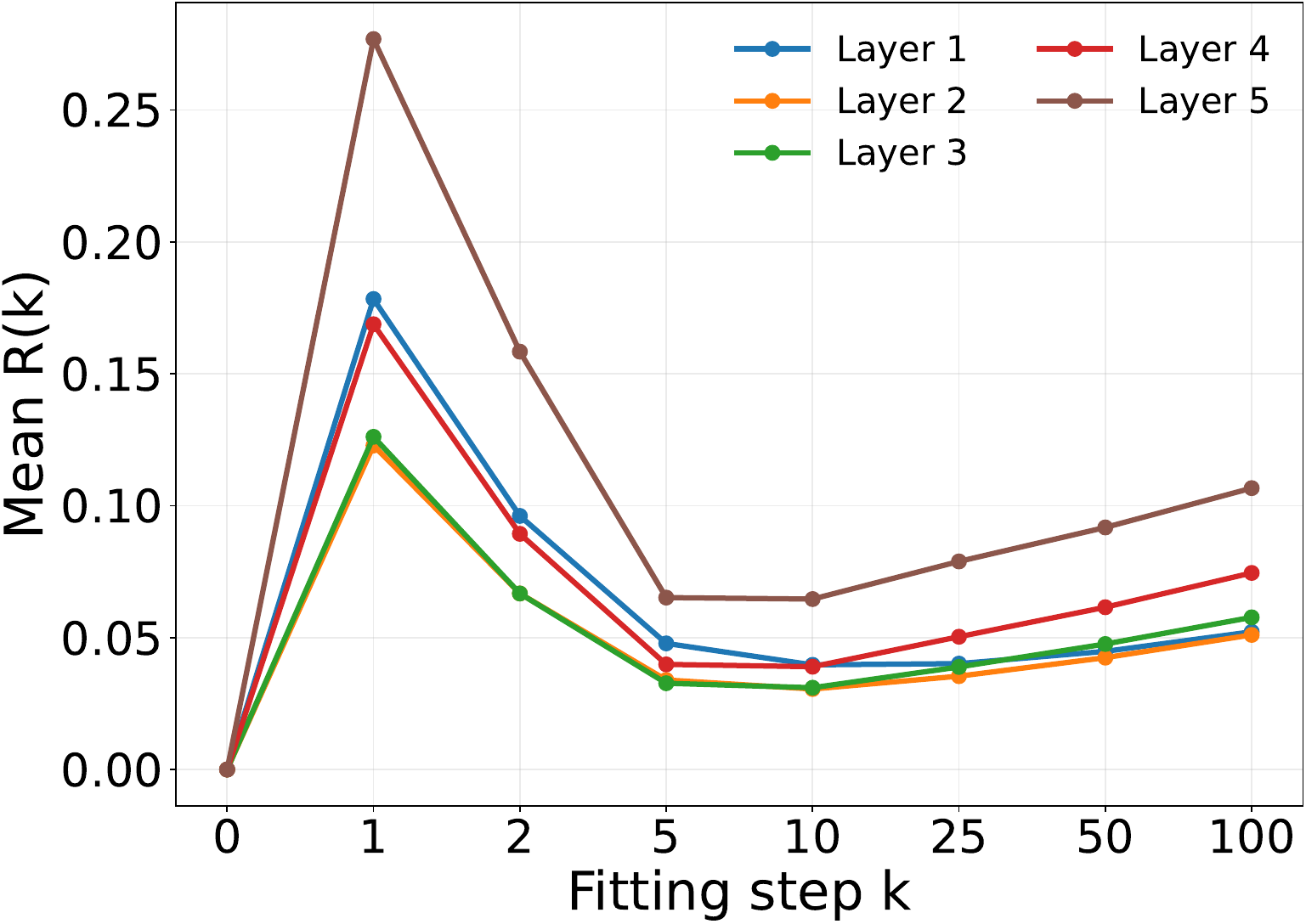}
        {(b) MetaSeg}
    \end{minipage}

    \vspace{0.5em}

    \caption{Mean normalized augmentation effect ($R(k)$ from eq.~\ref{eq:r}) for each INR layer during test-time INR fitting for HierINRSeg and MetaSeg (120k) trained without augmentation. In both models, the augmentation effect first spikes, then becomes relatively small after only a short period of fitting, indicating that as test-time INR adaptation progresses, effects of augmentation-induced variations decrease relative to the variations between different images. The same trend is observed for both HierINRSeg and MetaSeg trained with augmentation, as shown in Appendix Figure~\ref{fig:augmentation_effect_aug}.}
    \label{fig:augmentation_effect}
\end{figure}

Figure~\ref{fig:augmentation_effect} shows that, for both HierINRSeg and MetaSeg, the normalized augmentation effect first spikes at the beginning of test-time fitting, then quickly decreases and remains low until fitting completes at k=100 steps. Despite the fact that neither model was trained with augmentation, the spike is still quickly suppressed. This may explain the reason why standard train-time augmentation provides limited additional gains.

We also observe that deeper features are relatively more influenced by augmentations throughout the fitting process (matching our motivations for HierINRSeg). This is particularly relevant for MetaSeg, which uses only the fifth INR layer for segmentation. This corresponds well to our observation that MetaSeg is relatively more affected by augmentation, with an example shown in Appendix Table~\ref{tab:augmentation_effect_120k}. 

It is important to note that this trend does not mean that test-time fitting steps directly reduce the absolute effect of data augmentation. According to Appendix Figures~\ref{fig:raw_sim_0},~\ref{fig:raw_sim_1},~\ref{fig:raw_sim_2}, and~\ref{fig:raw_sim_3}, the absolute values of both augmentation-induced feature differences and cross-image feature differences increase as fitting progresses. However, the early drop in the relative ratio of augmentation effects still shows that the changes induced by augmentation, which are not part of the training, can be quickly suppressed by image-specific variation, which is part of the training. This suggests that, to fully exploit augmentation, rather than relying only on standard image-level perturbations, INR-based segmentation models may require more carefully designed strategies. This motivates future work on INR-specific augmentation methods, such as perturbations applied during fitting or augmentation to its feature-space representations.

\subsection{When Could INRs be Chosen?}

Overall, our experiments reveal distinct scaling patterns between INR-based and U-Net-based segmentation models, providing empirical insights into the settings where each model family may be favored. In the in-domain setting, the relative performance is mainly influenced by the parameter budget: INR-based models perform better at smaller parameter budgets, while U-Net-based models benefit more from increasing model capacity and become more competitive at larger budgets.

In the out-of-domain setting, the scaling behavior also depends on training augmentation. Without augmentation, INR-based models maintain stronger performance across most of our budget range. However, when augmentation becomes available, U-Nets obtain improved performance, narrowing or reversing the gap at medium and larger parameter budgets. This suggests that both parameter budget and available augmentation strategies should be considered when comparing their expected performance.

Under broader 3D spatial coverage, INR-based models can still perform well at very small parameter budgets, but their performance tends to decrease as spatial coverage increases. In contrast, U-Nets appear to be relatively less affected and may even benefit from greater spatial coverage. Therefore, the relative scaling behavior of INR- and U-Net-based models is also affected by the dataset's 3D spatial coverage and its associated increase in spatial variability.

\section{Ablation and Component Analysis}

\textbf{Disentangling our improvements from head capacity:} In the parameter-scaling experiments above, both the INR and the segmentation head are scaled for MetaSeg and HierINRSeg. To verify that our improvement is not simply due to a relatively larger segmentation head, we keep the INR size fixed and vary only the capacity of the MetaSeg segmentation head. The results are shown in Table~\ref{tab:metaseg_size_ablation}. All runs are conducted under the \textit{practical comparison} setting with augmentation and are repeated with three seeds. The default hidden dimension for all models is 128 (the size used for 2D/5-class in the MetaSeg codebase).

We increase the MetaSeg segmentation head capacity by varying both its width and depth, and also include two variants with smaller heads. We observe that head capacity has only a limited effect on performance: increasing or decreasing the head size does not substantially change either ID or OOD performance. In contrast, our method achieves substantially better performance than MetaSeg even when MetaSeg is equipped with much larger segmentation heads. This confirms that the improvement is not merely due to increased head capacity, but instead comes from incorporating features from additional INR layers.

\begin{table}[tb]
\centering
\caption{
Head-capacity ablation for MetaSeg with the INR kept fixed.
All models are trained on KKI, and Dice scores are averaged across three seeds.
Parameter counts are reported as INR + head. W denotes the width of the hidden layers of the segmentation head. D denotes the number of hidden layers. 
}
\label{tab:metaseg_size_ablation}
\small
\setlength{\tabcolsep}{4pt}
\begin{tabular}{llccc}
\toprule
\multirow{2}{*}{Head config} & \multirow{2}{*}{Params} & ID & \multicolumn{2}{c}{OOD} \\
\cmidrule(lr){3-3} \cmidrule(lr){4-5}
 & & KKI & Caltech & CMU \\
\midrule
W32-D1 & 67K + 4K   & 0.670 ± 0.001 & 0.562 ± 0.010 & 0.656 ± 0.013 \\
W64-D1 & 67K + 8K   & 0.676 ± 0.012 & 0.560 ± 0.007 & 0.652 ± 0.012 \\
W128-D1 (MetaSeg) & 67K + 17K  & 0.685 ± 0.005 & 0.575 ± 0.014 & 0.658 ± 0.005 \\
W256-D1 & 67K + 34K  & 0.689 ± 0.020 & 0.575 ± 0.026 & 0.659 ± 0.020 \\
W512-D1 & 67K + 69K  & 0.687 ± 0.009 & 0.575 ± 0.006 & 0.656 ± 0.006 \\
W512-D2 & 67K + 331K & 0.691 ± 0.019 & 0.582 ± 0.015 & 0.651 ± 0.016 \\
W512-D3 & 67K + 594K & 0.672 ± 0.005 & 0.565 ± 0.005 & 0.627 ± 0.022 \\
\midrule
HierINRSeg & 67K + 138K & \textbf{0.746} ± 0.003 & \textbf{0.683} ± 0.019 & \textbf{0.738} ± 0.007 \\
\bottomrule
\end{tabular}
\end{table}

\vspace{3pt} 
\noindent\textbf{Layer contribution to performance improvement:} We next examine how much each group of INR layers contributes to the final segmentation performance. To do this, we train variants of our segmentation head that use different subsets of INR features instead of all intermediate layers. All experiments are conducted under the \textit{practical comparison} setting and averaged over three seeds.

As shown in Table~\ref{tab:component_ablation}, performance generally improves as more INR layers are included. The five-layer variant performs the best on both OOD test sets and is only slightly below the four-layer variant on the ID test set. This suggests that using all intermediate layers is especially beneficial for cross-domain generalization. We also include a variant that uses only the first and last layers while excluding the middle layers (\textit{ours: excl. middle}). This variant performs worse than the five-layer version, showing that the middle layers provide useful information that cannot be recovered from the shallowest and deepest features alone. This supports our motivating analysis, where the middle layers were observed to preserve richer anatomical structure.

\begin{table}[tb]
\centering
\caption{
Component ablation of HierINRSeg where INR features from different sets of layers are used. All models are trained on KKI, and Dice scores are averaged across three seeds. Parameter counts are reported as INR + head.
}
\label{tab:component_ablation}
\small
\setlength{\tabcolsep}{4pt}
\begin{tabular}{llcccccccc}
\toprule
\multirow{2}{*}{Model} & \multirow{2}{*}{Params} & \multicolumn{5}{c}{Selected layers} & ID & \multicolumn{2}{c}{OOD} \\
\cmidrule(lr){3-7} \cmidrule(lr){8-8} \cmidrule(lr){9-10}
 & & 1 & 2 & 3 & 4 & 5 & KKI & Caltech & CMU \\
\midrule
 MetaSeg & 67K+17K & &  &  &  & \checkmark & 0.685 ± 0.005 & 0.575 ± 0.014 & 0.658 ± 0.005 \\
 MetaSeg: large & 67K + 594K & &  &  &  & \checkmark & 0.672 ± 0.005 & 0.565 ± 0.005 & 0.627 ± 0.022  \\
\midrule
HierINRSeg: 1 layer &67K+3K & &  &  &  & \checkmark & 0.700 ± 0.014 & 0.586 ± 0.005 & 0.690 ± 0.005 \\
HierINRSeg: 2 layers & 67K+36K & & &  & \checkmark & \checkmark & 0.727 ± 0.023 & 0.636 ± 0.010 & 0.716 ± 0.010 \\
HierINRSeg: 3 layers &67K+70K & &  & \checkmark & \checkmark & \checkmark & 0.731 ± 0.006 & 0.638 ± 0.003 & 0.721 ± 0.010 \\
HierINRSeg: excl. middle & 67K+36K& \checkmark &  &  &  & \checkmark & 0.730 ± 0.013 & 0.632 ± 0.041 & 0.712 ± 0.009 \\
HierINRSeg: 4 layers & 67K+104K& & \checkmark & \checkmark & \checkmark & \checkmark & \textbf{0.749} ± 0.016 & 0.670 ± 0.001 & 0.731 ± 0.006 \\
HierINRSeg: 5 layers (all) &67K+138K & \checkmark & \checkmark & \checkmark & \checkmark & \checkmark & 0.746 ± 0.003 & \textbf{0.683} ± 0.019 & \textbf{0.738} ± 0.007 \\
\bottomrule
\end{tabular}
\end{table}

\section{Limitations}
The current augmentation strategy is limited to intensity-based transformations. Future work should incorporate spatial perturbations, which MetaSeg also identified as a limitation for INRs. In addition, our results show that INR-based models become more sensitive as 3D spatial coverage increases, suggesting that their performance on datasets with greater structural variability requires further investigation. Our experiments were conducted for brain MRI segmentation on one in-domain site and two out-of-domain sites from ABIDE I to provide a controlled evaluation setting. Further validation of these findings on broader and more diverse datasets will be needed in future work.

\section{Conclusion}
In this work, we studied INR-based semantic segmentation for brain MRI under low-parameter and cross-domain settings. We showed that complementary segmentation-relevant information is distributed across multiple INR layers, motivating HierINRSeg, a new INR-based segmentation architecture that hierarchically fuses multi-layer INR features. We conducted comprehensive experiments comparing INR-based models and conventional U-Net-based baselines across parameter budgets and augmentation settings, showing that HierINRSeg consistently improves over MetaSeg. By observing performance patterns across different budgets and training conditions, we further identified distinct scaling behaviors between the two model families: INR-based models perform well when the parameter or augmentation budget is limited, while U-Net-based models benefit more from increased model capacity and data augmentation, becoming more competitive at larger parameter budgets and under broader 3D spatial coverage. Our work lays the foundation for future work on INR-specific augmentation, domain adaptation strategies, and more diverse segmentation settings.

\FloatBarrier

\bibliography{main}
\bibliographystyle{tmlr}

\appendix

\section{Label map generation}
The original label maps were derived from FreeSurfer segmentations and converted into a reduced tissue-label representation for 2D segmentation. Each voxel in the original volume was assigned to one of six output classes: 0 for background or ignored structures, 1 for cerebral white matter, 2 for cerebral cortical gray matter, 3 for CSF and ventricular regions, 4 for subcortical gray matter, and 5 for cerebellum.

Cerebral white matter included FreeSurfer labels [2, 41, 77, 251, 252, 253, 254, 255]. Cerebral cortical gray matter included labels 3 and 42. CSF and ventricular regions included labels [4, 5, 14, 15, 24, 43, 44, 72, 31, 63]. Subcortical gray matter included labels [10, 49, 11, 50, 12, 51, 13, 52, 17, 53, 18, 54, 26, 58, 28, 60, 16, 85], corresponding to deep gray matter and related non-cortical brain structures. Cerebellar white matter and cortex were merged into a single cerebellum class using labels [7, 46, 8, 47]. Vessel labels 30 and 62, along with unlabeled or ignored regions, were mapped to the background.

\section{Model Sizes}
Table~\ref{tab:appendix-unet-sizes} lists the U-Net model sizes used in our experiments. The feature count tuple specifies the number of channels used at each resolution level of the network encoder and bottleneck. For example, the tuple $(4, 7, 12, 26, 52)$ means that the encoder uses 4, 7, 12, and 26 feature channels and the bottleneck uses 52 channels. The decoder uses the same structure in reverse. Larger tuples correspond to wider U-Nets with higher parameter counts. The model sizes for UNeXt are listed the same way in Table~\ref{tab:unext_model_sizes}.

All INRs use the same backbone depth: 5 hidden SIREN layers, input dimension 2, and output dimension 1. Model size is controlled by a hidden width held constant across hidden layers within a given model. Tables~\ref{tab:appendix-inr-orig-sizes} and~\ref{tab:appendix-inr-gradual-sizes} report the sizes for MetaSeg and HierINRSeg. To include the original default MetaSeg configuration, we ensure that its default hidden width of 128 is included in both INR sweeps. This corresponds to $\sim$84K parameters for MetaSeg and $\sim$205K for HierINRSeg, slightly above the nominal 80K and 200K targets. These deviations are tiny relative to the budgets (4.8\% and 2.6\%, respectively) and are not expected to materially affect the parameter-matched comparisons.

\begin{table}[hbtp]
\centering
\small
\caption{U-Net model sizes.}
\label{tab:appendix-unet-sizes}
\begin{tabular}{lll}
\toprule
Target Size & Feature counts & Parameters \\
\midrule
5k    & $(1,1,4,6,13)$      & 5,037 \\
10k   & $(1,2,5,9,18)$      & 10,005 \\
20k   & $(2,4,8,13,23)$     & 19,999 \\
40k   & $(2,5,10,18,36)$    & 40,011 \\
70k   & $(3,6,12,24,49)$    & 70,003 \\
80k   & $(4,7,12,26,52)$    & 79,999 \\
120k  & $(4,9,15,32,63)$    & 119,706 \\
200k  & $(5,11,20,42,80)$   & 200,038 \\
500k  & $(8,16,32,66,128)$  & 499,992 \\
700k  & $(9,19,39,75,157)$  & 700,012 \\
1m    & $(12,23,47,91,184)$ & 999,965 \\
\bottomrule
\end{tabular}
\end{table}

\begin{table}[t]
\centering
\small
\caption{Metaseg model sizes.}
\label{tab:appendix-inr-orig-sizes}
\begin{tabular}{lrr}
\toprule
Target size & $d$ (layer width)& Parameters \\
\midrule
5k   & 30  & 4,957 \\
10k  & 43  & 9,897 \\
20k  & 62  & 20,157 \\
40k  & 88  & 40,047 \\
70k  & 117 & 70,207 \\
80k  & 128 & 83,847 \\
120k & 153 & 119,347 \\
200k & 199 & 200,997 \\
500k & 315 & 500,857 \\
700k & 373 & 701,247 \\
1m   & 446 & 1,001,277 \\
\bottomrule
\end{tabular}
\end{table}

\begin{table}[t]
\centering
\small
\caption{HierINRSeg model sizes.}
\label{tab:appendix-inr-gradual-sizes}
\begin{tabular}{lrr}
\toprule
Target size & $d$ (layer width) & Parameters \\
\midrule
5k   & 18  & 5,101 \\
10k  & 26  & 9,861 \\
20k  & 38  & 19,881 \\
40k  & 55  & 39,992 \\
70k  & 74  & 70,677 \\
80k  & 79  & 80,192 \\
120k & 97  & 119,414 \\
200k & 128 & 205,191 \\
500k & 201 & 498,286 \\
700k & 239 & 701,472 \\
1m   & 286 & 1,000,721 \\
\bottomrule
\end{tabular}
\end{table}

\begin{table}[t]
\centering
\caption{UNeXT model sizes. Feature counts specify the channel widths
at the five stages.}
\label{tab:unext_model_sizes}
\begin{tabular}{lcc}
\toprule
Target size & Feature counts & Parameters \\
\midrule
5k & $(1, 3, 5, 6, 21)$ & $5{,}000$ \\
10k & $(3, 3, 5, 12, 25)$ & $10{,}000$ \\
20k & $(2, 3, 5, 20, 34)$ & $20{,}000$ \\
40k & $(1, 6, 12, 25, 51)$ & $40{,}000$ \\
70k & $(3, 7, 18, 31, 71)$ & $70{,}000$ \\
80k & $(6, 11, 17, 34, 74)$ & $80{,}000$ \\
120k & $(5, 14, 21, 46, 83)$ & $120{,}000$ \\
200k & $(7, 13, 27, 55, 119)$ & $200{,}000$ \\
500k & $(16, 26, 45, 88, 182)$ & $500{,}000$ \\
700k & $(13, 29, 51, 104, 221)$ & $700{,}000$ \\
1m & $(13, 35, 66, 127, 255)$ & $1{,}000{,}000$ \\
\bottomrule
\end{tabular}
\end{table}

\section{Implementation Details}
\label{sec:impl_details}
The training for MetaSeg and HierINRSeg follows the two-stage MetaSeg procedure, with a first meta-training stage for the INR and segmentation head, followed by a segmentation-head training stage using features extracted from fitted INRs. The number of outer-loop iterations in the first stage is capped at 5000, with the INR selected from the iteration with the highest validation mean Dice (average across all foreground classes and validation images). For the second step, the segmentation head is trained for up to 4000 epochs, with early stopping after 5 consecutive non-improving validation rounds. The head with the lowest validation loss is selected. MetaSeg and HierINRSeg share the same set of training hyperparameters listed in Table~\ref{tab:training_hyperparameters}. 

For the U-Net baselines, we used a standard 2D U-Net with five resolution levels, two Conv-BN-ReLU blocks per stage, and channel widths that varied according to the preset parameter budget. The network was trained for 100 epochs with Adam (lr=1e-3, batch size=8) using cross-entropy loss, with the best validation mean Dice checkpoint used for evaluation.

\begin{table}[h]
\centering
\begin{tabular}{ll}
\hline
\textbf{Hyperparameter} & \textbf{Value} \\
\hline
Step 1 inner-loop fitting steps & 2 \\
Validation/test time INR fitting steps & 100 \\
Validation frequency & Every 50 iterations \\
Focal loss: & gamma=1, no class weights \\
INR learning rate & $1 \times 10^{-4}$ \\
Classifier learning rate & $5 \times 10^{-5}$ \\
INR number of hidden layers & 5 \\
Step 2 augmentation refresh frequency & Every 20 epochs \\
Step 2 validation frequency & Every 50 epochs \\
\hline
\end{tabular}
\caption{Remaining training and validation hyperparameters.}
\label{tab:training_hyperparameters}
\end{table}

We report ordinary Dice without background, averaging foreground classes per subject and treating classes absent in both prediction and ground truth as 1 and classes present on only one side as 0.

\textbf{nnU-Net baseline.} We use the official nnU-Net library (v2) in its 2D configuration. Each KKI training slice is padded to a square and resampled to 160×160, then converted into the format expected by the library. We let nnU-Net plan the architecture and patch size from the data fingerprint; on KKI it selects a six-stage PlainConvUNet (features 32, 64, 128, 256, 512, 512 with 3×3 InstanceNorm and LeakyReLU convolutional blocks) trained on 128×160 patches, totaling approximately 20.6 M parameters. The dataset's train and validation split is injected in place of nnU-Net's automatic five-fold cross-validation, so a single fold is trained on the dataset's exact partition. Training follows SGD with Nesterov momentum 0.99, polynomial learning-rate decay (initial 0.01, power 0.9), and a combined Dice and cross-entropy loss with the background class excluded. We use the framework's default on-the-fly augmentation, which includes rotation, scaling, gamma, brightness and contrast, additive and multiplicative Gaussian noise, low-resolution simulation, and Gaussian blur. The only departure from defaults is disabling left and right mirroring, since brain anatomy is lateralized. Three independent seeds are trained for 250 epochs each.

\textbf{TransUNet baseline.} We use the public 2D TransUNet implementation with the hybrid R50-ViT-B/16 configuration. Each KKI training slice is zero-padded to a square and resized to 160$\times$160; image slices use bilinear interpolation and label maps use nearest-neighbor interpolation. The single-channel MRI input is internally replicated to three channels to match the ImageNet-pretrained hybrid encoder. The network uses a ResNetV2 hybrid backbone with a ViT-B transformer encoder (12 layers, 12 attention heads, hidden size 768, MLP size 3072) operating on a 10$\times$10 token grid, followed by the standard TransUNet convolutional decoder with channels 256, 128, 64 and 16 and three skip connections. The final segmentation head predicts six classes, totalling approximately 105.2 M trainable parameters. We use a batch size of 24 and SGD with momentum 0.9, weight decay $10^{-4}$, polynomial learning-rate decay (initial 0.01, power 0.9), and an equally weighted sum of cross-entropy and multiclass Dice loss. Its original data augmentation is replaced by the shared augmentation used for the INR/U-Net baselines. Three independent seeds are trained for 150 epochs each.

\textbf{UNeXT baseline.} We implemented the original 2D UNeXT architecture with one input channel and six output classes, scaling channel widths to match each parameter budget. Models were trained for 500 epochs using Adam (learning rate $10^{-3}$), cross-entropy loss, batch size 8, and \(160 \times 160\) images. We used the HierINRSeg dataloader and augmentation scheme exclusively, with augmentation enabled only during training. The best checkpoint was selected by validation Dice. Results are averaged over three seeds.

\textbf{Data Augmentation for Practical Comparison.}
Each image undergoes a random
composed set of standard intensity augmentation with probability $p_{augment}=0.9$. This includes gamma transformation with $g \sim \mathcal{U}(0.8,1.2)$ and $p_{gamma}=0.5$, contrast scaling with $c \sim \mathcal{U}(0.8,1.2)$ and $p_{scale}=0.5$, intensity shift with $b \sim \mathcal{U}(-0.05,0.05)$ and $p_{shift}=0.5$, and Gaussian noise with $\sigma \sim \mathcal{U}(0,0.01)$ and $p_{gaussian}=0.25$.

\textbf{Sensitivity to 3D spatial coverage.} For the additional experiments at 20\%, 40\%, 60\%, and 80\% spatial coverage, with the subject-wise splits preserved, for each spatial-coverage setting, 25 slices are sampled equidistantly per subject from the corresponding central region, keeping the number of samples fixed while progressively increasing spatial coverage. This yields 825/175/200 slices for the training/validation/test sets, respectively. During training, to expedite model selection, the validation set is randomly subsampled to 100 slices.

\section{UMAP Hyperparameters and Sensitivity}
\begin{figure}[tb]
    \centering
    \includegraphics[width=0.9\linewidth]{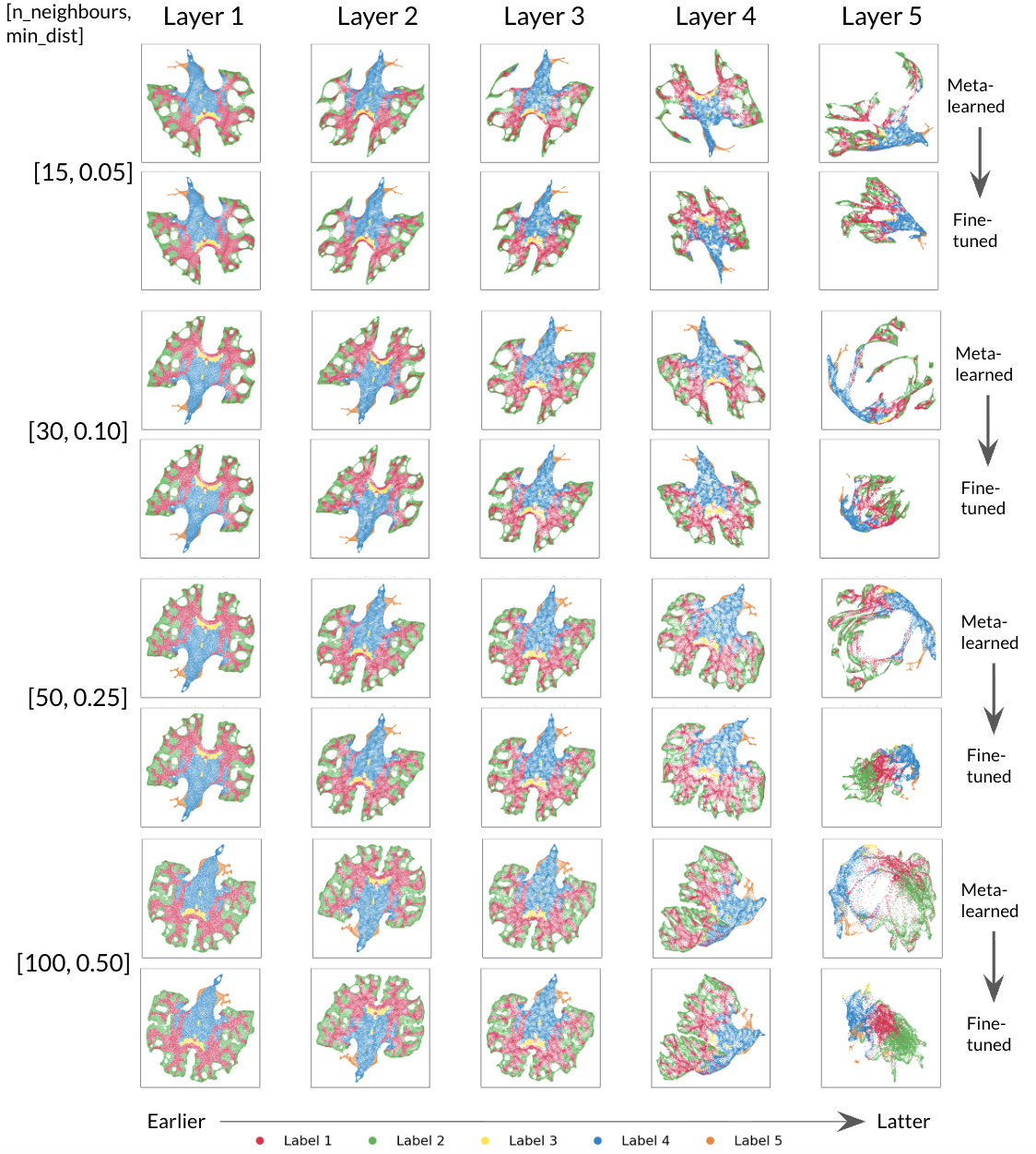}
    \caption{UMAP parameter sweep for visualizations shown in Figure~\ref{fig:motiv0}.}
    \label{fig:umap_sweep_1}
\end{figure}

\begin{figure}[tb]
    \centering
    \includegraphics[width=0.9\linewidth]{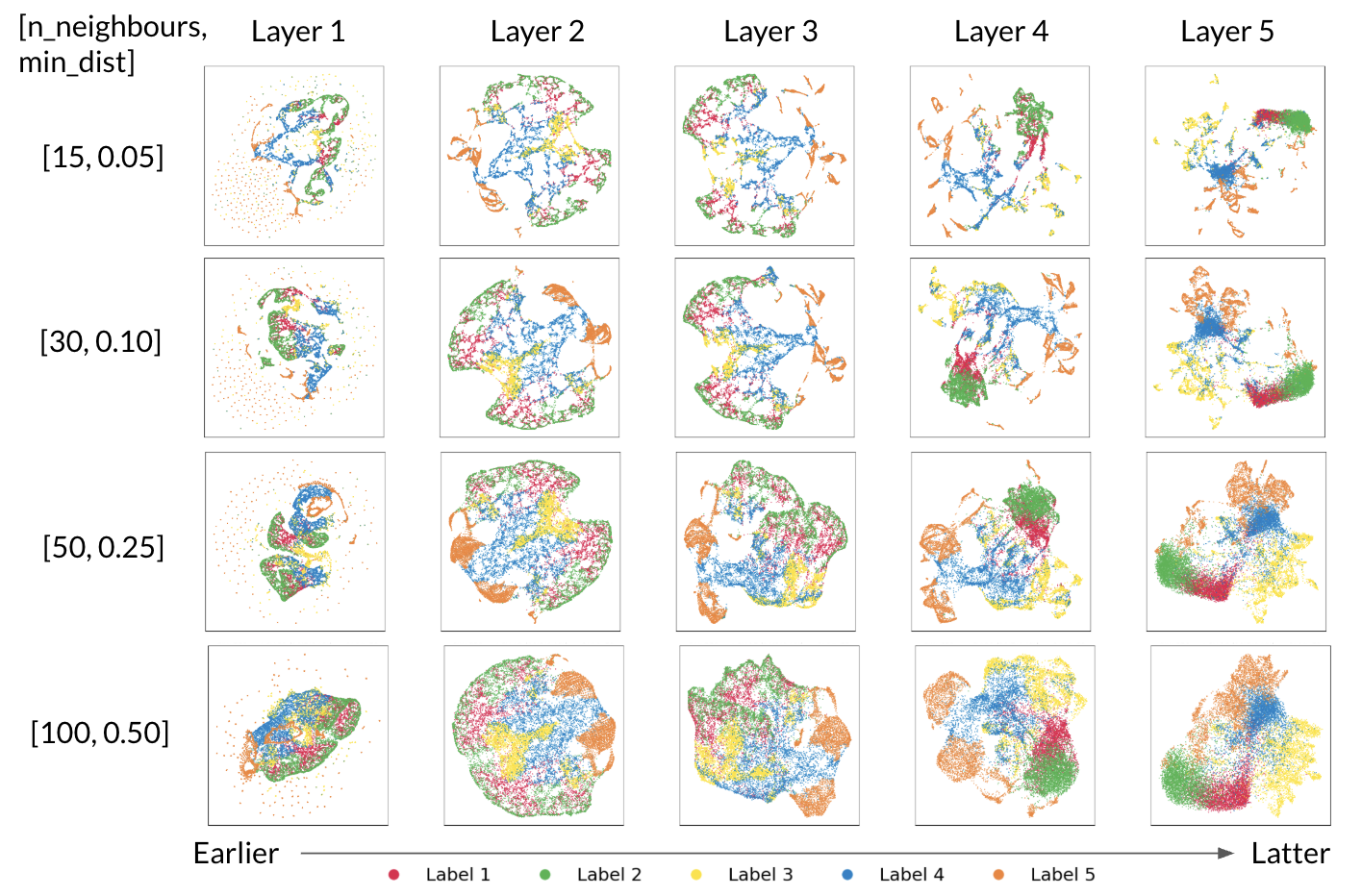}
    \caption{UMAP parameter sweep for visualizations shown in Figure~\ref{fig:motiv1}.}
    \label{fig:umap_sweep_2}
\end{figure}

The visualizations in Figures~\ref{fig:motiv0} and~\ref{fig:motiv1} both used UMAP with hyperparameters \texttt{\{n\_neighbors=30, min\_dist=0.1\}}. Since UMAP outputs are known to be sensitive to hyperparameters, we extend both visualizations to a parameter sweep of \texttt{[\{15, 0.05\}, \{30, 0.10\}, \{50, 0.25\}, \{100, 0.50\}]}. As shown in Figures~\ref{fig:umap_sweep_1} and~\ref{fig:umap_sweep_2}, despite differing plot layouts, the reported layer-wise trends were surprisingly persistent in all parameter combinations. In Figure~\ref{fig:umap_sweep_1}, the spatial layouts of initial layers remain consistent before/after fine-tuning, while spatial layouts are significantly altered in the final layer. In Figure~\ref{fig:umap_sweep_2}, no clear structures are displayed in the first layer. Spatial structures resembling the brain begin to appear starting from the second layer and gradually transition into the label-discriminative clusters present in the final layer. 

\section{Additional Analysis}
\label{appn:additional_anal}

Here, we provide additional analyses to complement the results in the main paper. 

\begin{figure}[tb]
    \centering

    \noindent
    \begin{minipage}[t]{0.4\linewidth}
        \centering
        \includegraphics[width=\linewidth]{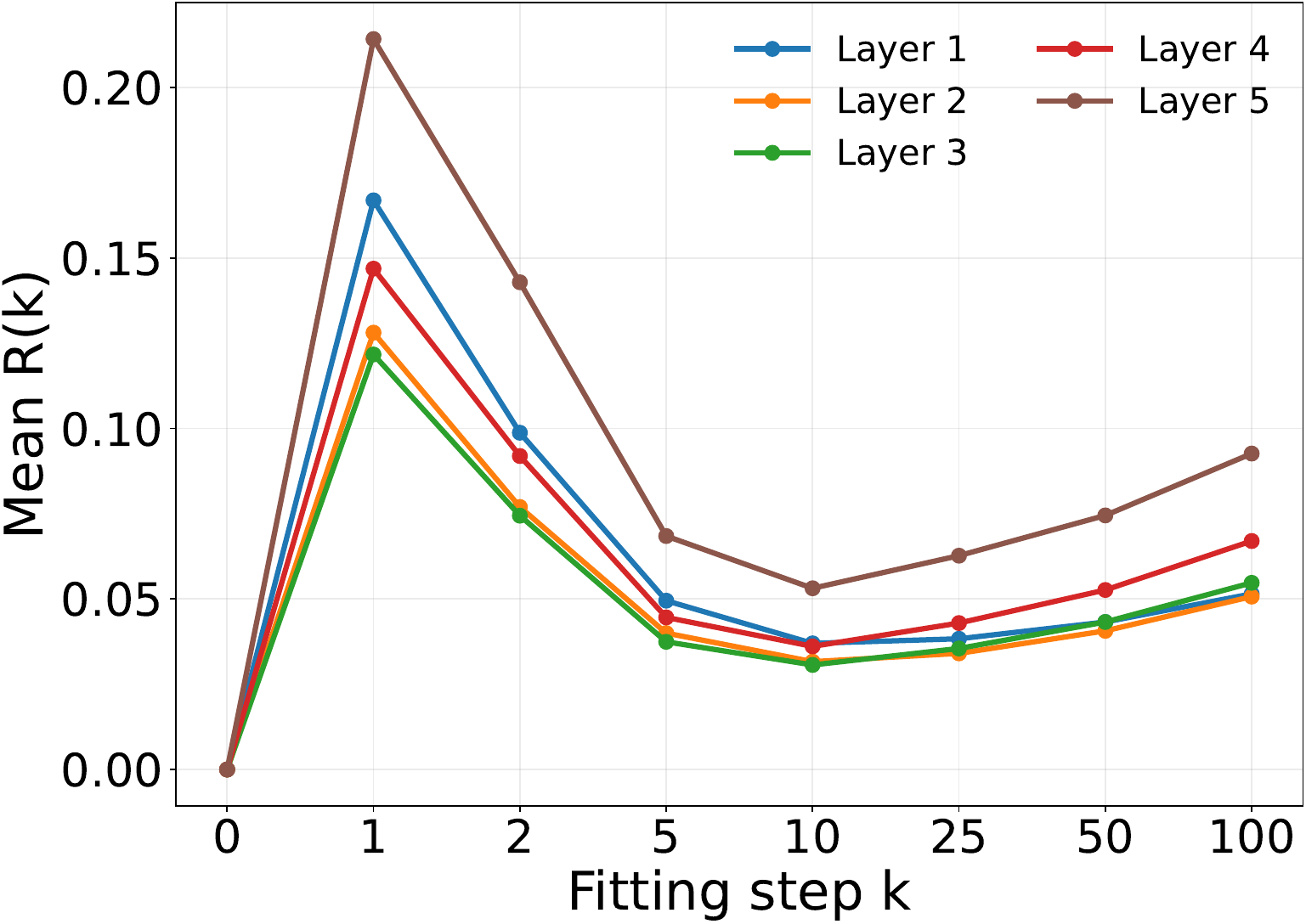}
        {Model: (a) HierINRSeg}
    \end{minipage} 
    \hspace{3em}
    \begin{minipage}[t]{0.4\linewidth}
        \centering
        \includegraphics[width=\linewidth]{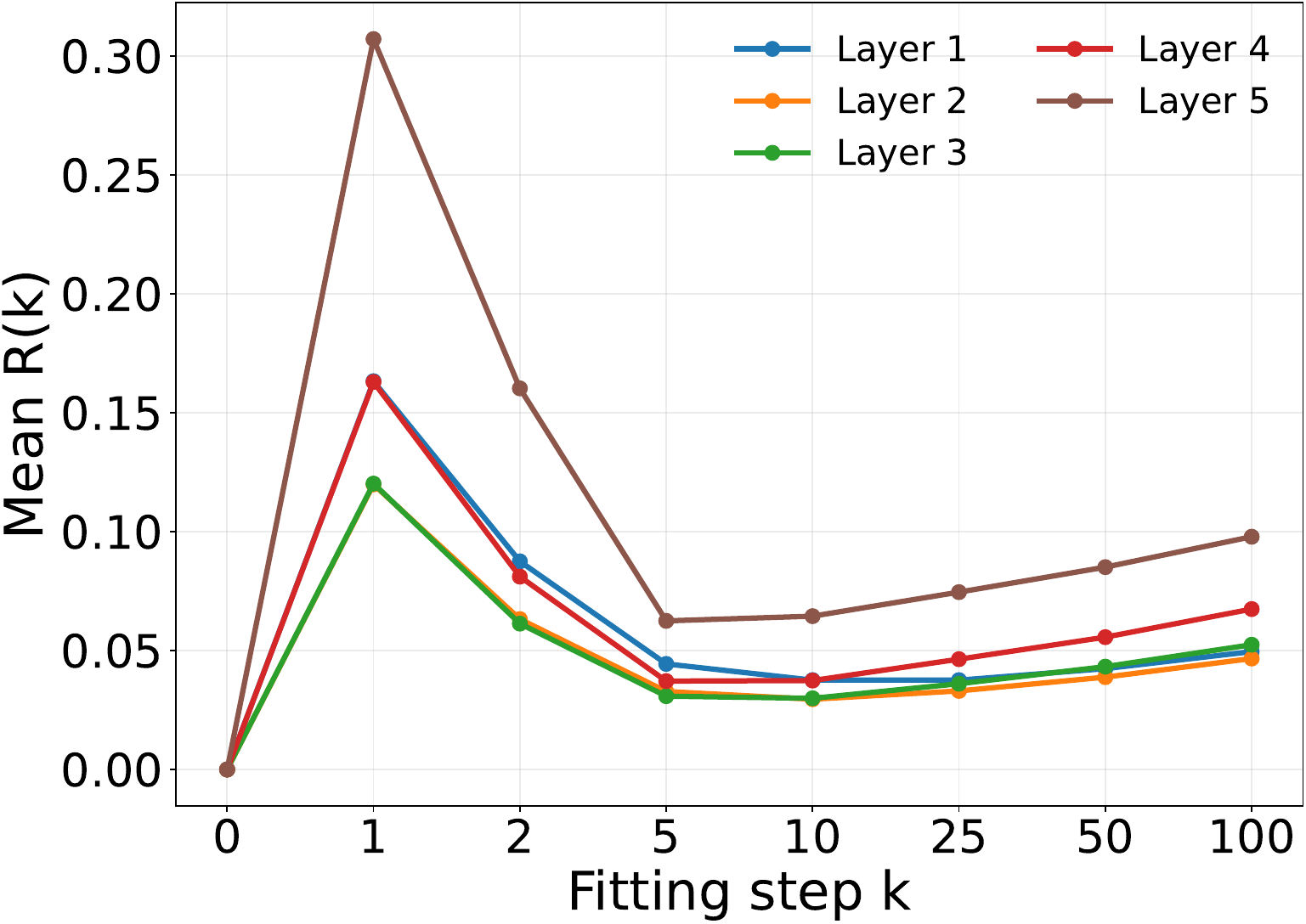}
        {(b) MetaSeg}
    \end{minipage}

    \vspace{0.5em}

    \caption{Normalized augmentation effect during test-time INR fitting for HierINRSeg and MetaSeg (120k) trained \textbf{with} augmentation. In both models, the relative augmentation effect becomes small after only a short period of fitting, showing the same trend as Figure~\ref{fig:augmentation_effect}.}
    \label{fig:augmentation_effect_aug}
\end{figure}

\begin{figure}[htbp]
    \centering
    \includegraphics[width=0.9\linewidth]{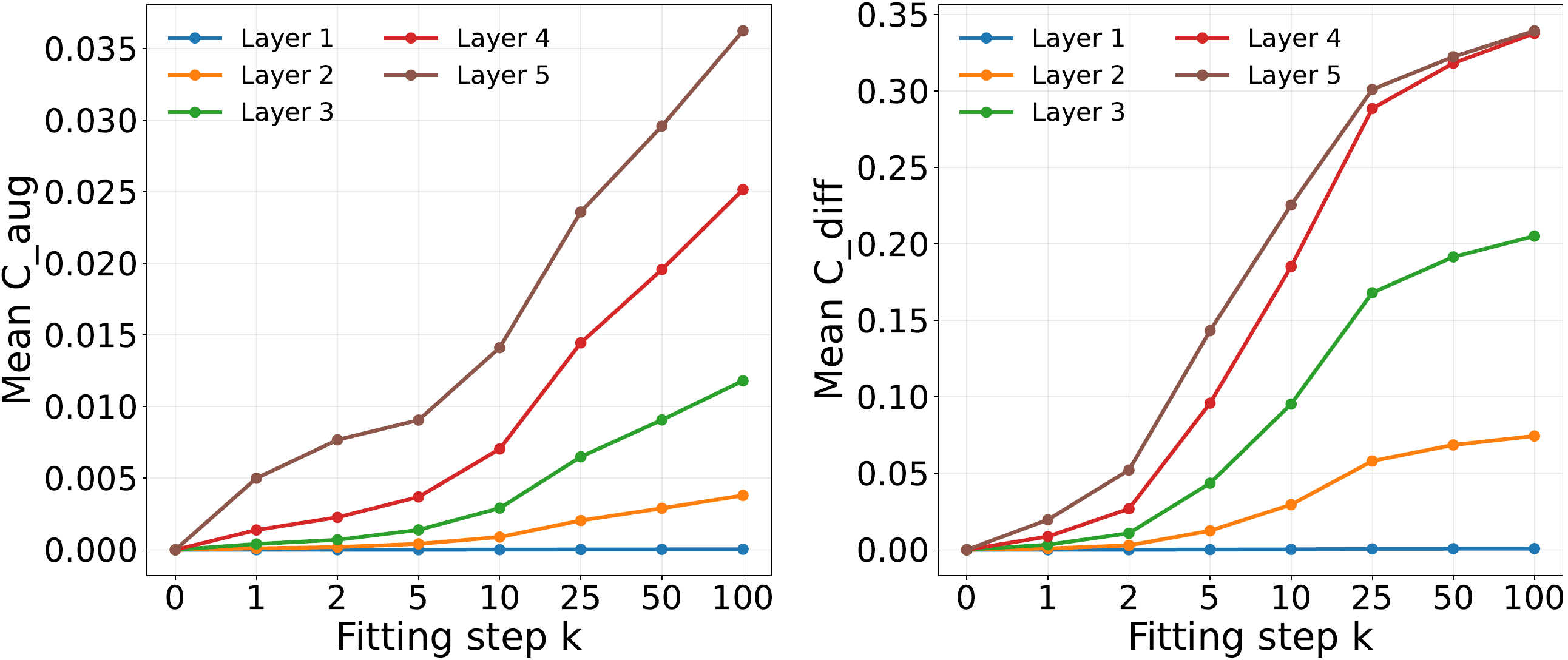}
    \caption{The absolute extents of feature changes caused by augmentation (left) and image variation (right) for MetaSeg trained without augmentation}
    \label{fig:raw_sim_0}
\end{figure}

\begin{figure}[htbp]
    \centering
    \includegraphics[width=0.9\linewidth]{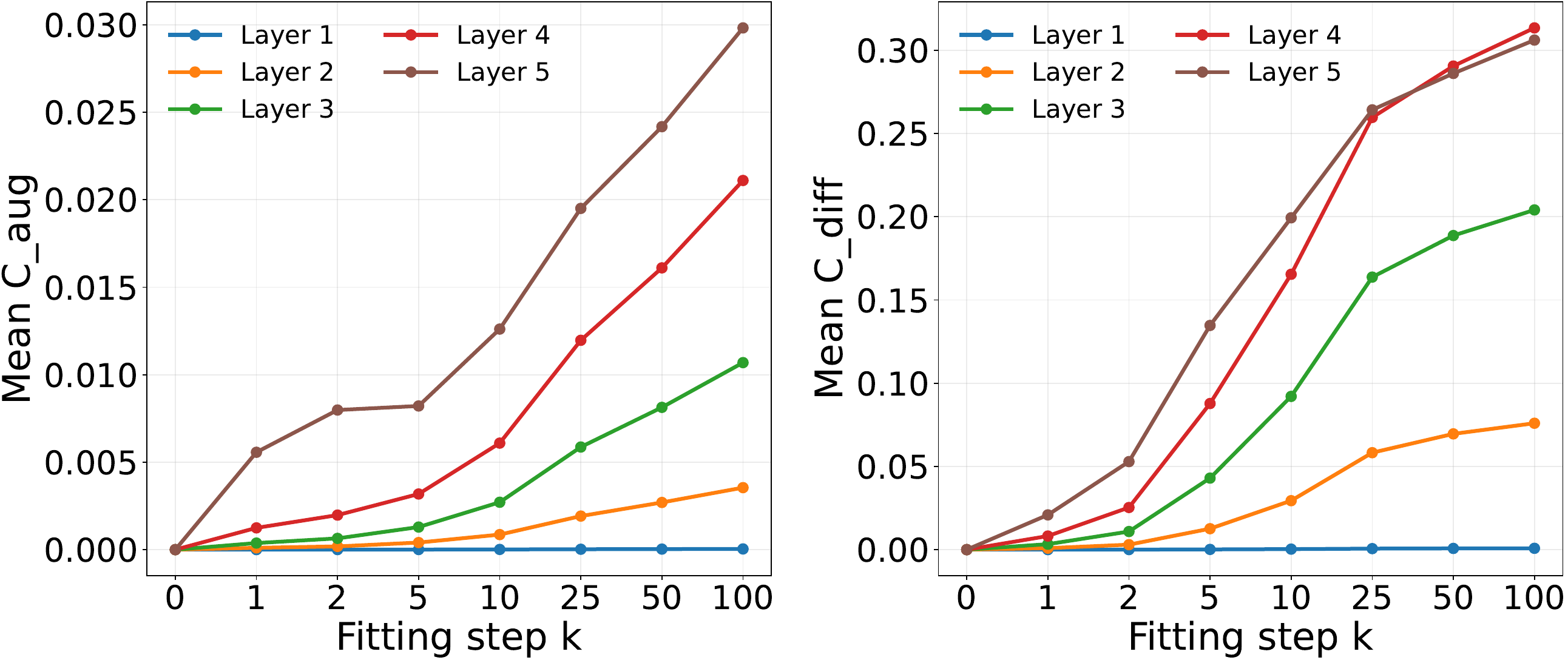}
    \caption{The absolute extents of feature changes caused by augmentation (left) and image variation (right) for MetaSeg trained with augmentation}
    \label{fig:raw_sim_1}
\end{figure}

\begin{figure}[htbp]
    \centering
    \includegraphics[width=0.9\linewidth]{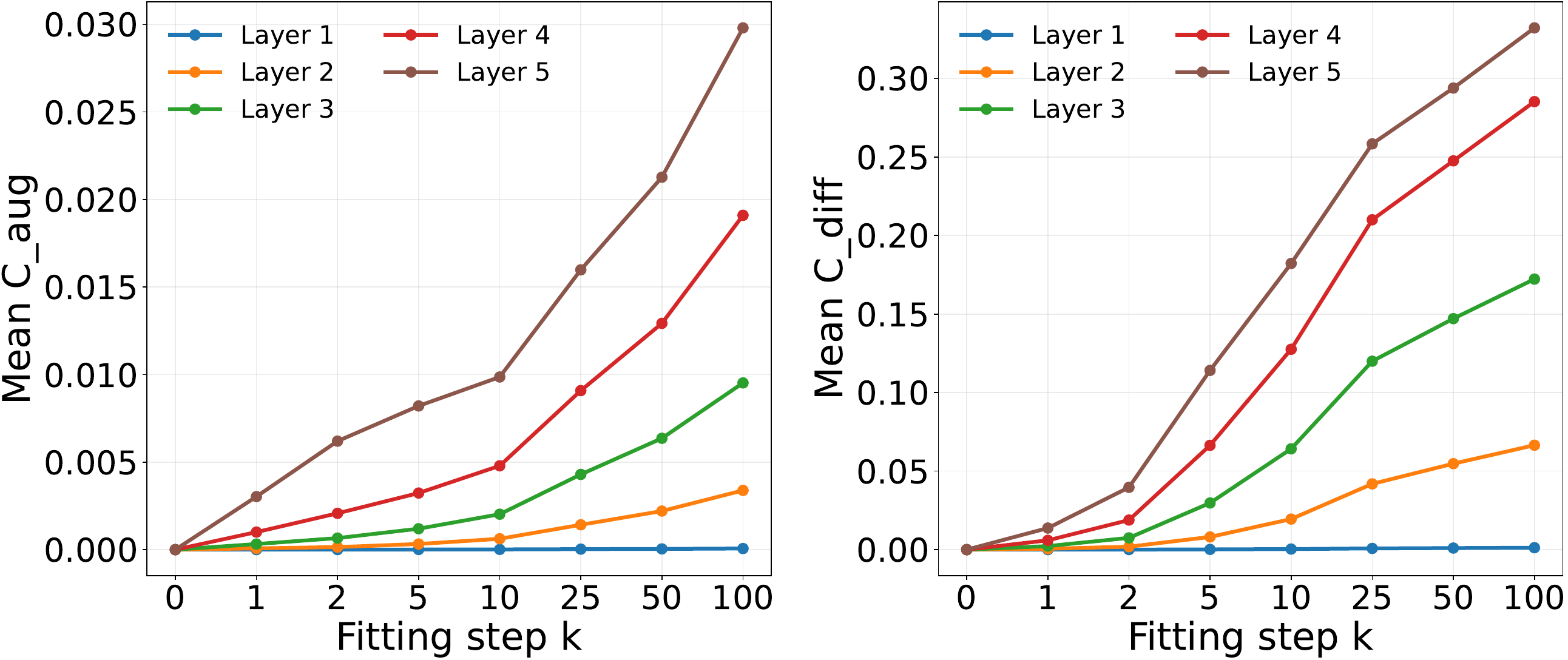}
    \caption{The absolute extents of feature changes caused by augmentation (left) and image variation (right) for HierINRSeg trained without augmentation}
    \label{fig:raw_sim_2}
\end{figure}

\begin{figure}[tb]
    \centering
    \includegraphics[width=0.9\linewidth]{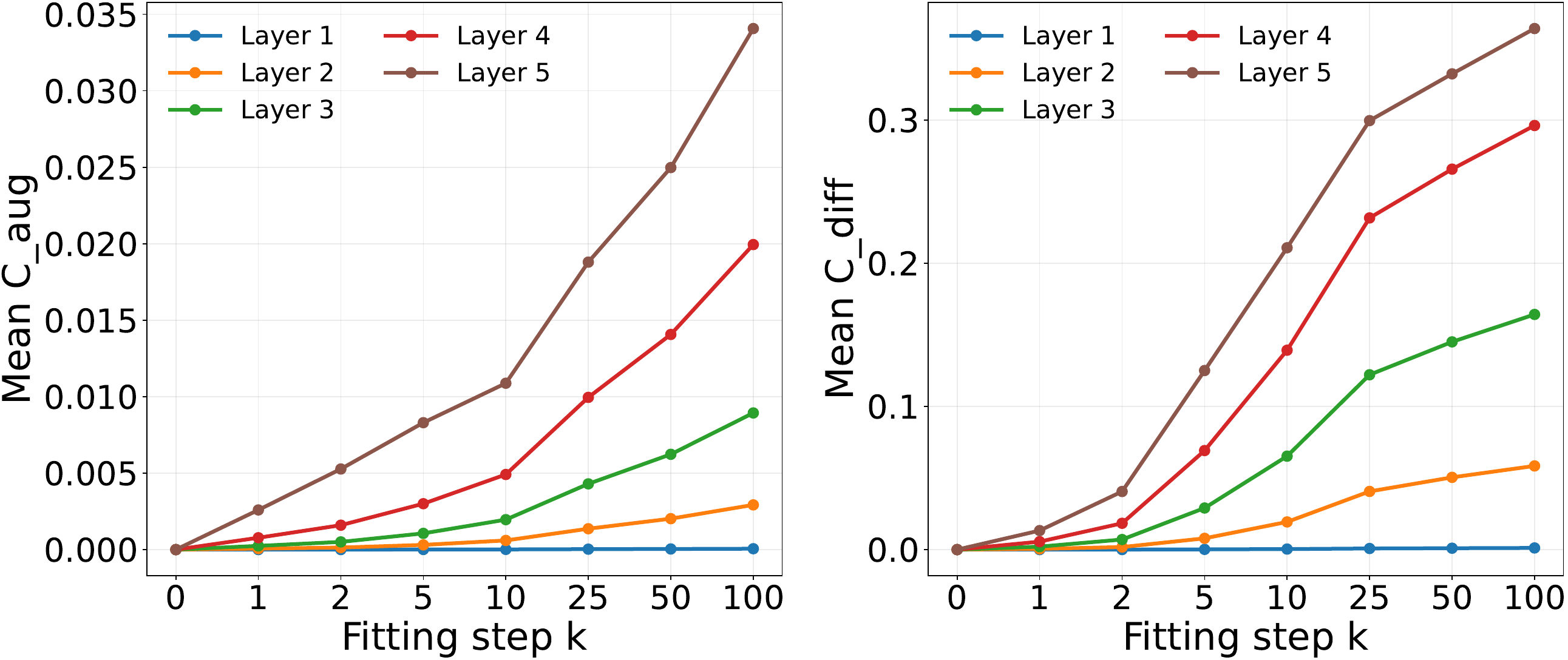}
    \caption{The absolute extents of feature changes caused by augmentation (left) and image variation (right) for HierINRSeg trained with augmentation}
    \label{fig:raw_sim_3}
\end{figure}

\textbf{Supplementing Section~\ref{sec:aug_sens_anal}} of the main paper, Figure~\ref{fig:augmentation_effect_aug} presents the augmentation-effect analysis for HierINRSeg and MetaSeg when both models are trained \textbf{with} augmentation. The normalized augmentation effect follows a trend that is largely consistent with the non-augmented setting reported in the main paper: as inference-time fitting progresses, the relative (normalized) effect of augmentation quickly decreases and remains low. This further supports our claim that inference-time INR fitting reduces model sensitivity to input perturbations (relative to between-image variations). Figures~\ref{fig:raw_sim_0},~\ref{fig:raw_sim_1},~\ref{fig:raw_sim_2}, and~\ref{fig:raw_sim_3} directly plot the $1 - S_l(I_k, A(I)_k)$ and $1 - S_l(I_k, I'_k)$ terms specified in Section~\ref{sec:aug_sens_anal}, which characterize the absolute extent of feature change versus the number of fitting steps. $1 - S_l(I_k, A(I)_k)$ is denoted as $C_{aug}$ and $1 - S_l(I_k, I'_k)$ is denoted as $C_{diff}$. The figures illustrate that the feature difference always increases as the number of fitting steps increases, though the pattern differs between augmentation-induced changes and between-image variations.

\begin{table}[hb]
\centering
\caption{Caltech (OOD) performance under increasing 3D spatial coverage. Slices are sampled from progressively wider central regions of each volume (5--80\%). Results are reported at representative 10K and 200K parameter budgets.}
\label{tab:spatial_variability_caltech}
\small
\setlength{\tabcolsep}{5pt}
\begin{tabular}{llccccc}
\toprule
\multirow{2}{*}{Budget} & \multirow{2}{*}{Model} 
& \multicolumn{5}{c}{Performance (Dice) on each percentage} \\
\cmidrule(lr){3-7}
& & 5\% & 20\% & 40\% & 60\% & 80\% \\
\midrule
\multirow{3}{*}{10K}
& MetaSeg    & 0.542 ± 0.016 & 0.518 ± 0.027 & 0.491 ± 0.021 & 0.432 ± 0.048 & 0.426 ± 0.012\\
& HierINRSeg &\textbf{ 0.596} ± 0.025 & \textbf{0.610} ± 0.009 & \textbf{0.587}± 0.009 & \textbf{0.534} ± 0.014 & 0.555 ± 0.017 \\
& U-Net      & 0.461 ± 0.027 & 0.419 ± 0.074 & 0.404 ± 0.043 & 0.467 ± 0.095 & \textbf{0.583} ± 0.011 \\
\midrule
\multirow{3}{*}{200K}
& MetaSeg    & 0.601 ± 0.017 & 0.583 ± 0.022 & 0.522 ± 0.008 & 0.406 ± 0.055 & 0.415 ± 0.040 \\
& HierINRSeg & 0.683 ± 0.019 & 0.699 ± 0.002 & 0.663 ± 0.004 & 0.610 ± 0.018 & 0.575 ± 0.015 \\
& U-Net      & \textbf{0.724} ± 0.018 & \textbf{0.719 }± 0.054 & \textbf{0.752} ± 0.009 & \textbf{0.720} ± 0.026 & \textbf{0.746} ± 0.042 \\
\bottomrule
\end{tabular}
\end{table}

\begin{table}[hb]
\centering
\caption{CMU (OOD) performance under increasing 3D spatial coverage. Slices are sampled from progressively wider central regions of each volume (5--80\%). Results are reported at representative 10K and 200K parameter budgets.}
\label{tab:spatial_variability_cmu}
\small
\setlength{\tabcolsep}{5pt}
\begin{tabular}{llccccc}
\toprule
\multirow{2}{*}{Budget} & \multirow{2}{*}{Model} 
& \multicolumn{5}{c}{Performance (Dice) on each percentage} \\
\cmidrule(lr){3-7}
& & 5\% & 20\% & 40\% & 60\% & 80\% \\
\midrule
\multirow{3}{*}{10K}
& MetaSeg    & 0.624 ± 0.010 & 0.560 ± 0.030 & 0.536 ± 0.029 & 0.475 ± 0.057 & 0.458 ± 0.013 \\
& HierINRSeg &\textbf{ 0.661} ± 0.006 & \textbf{0.660} ± 0.010 & \textbf{0.624 }± 0.002 & \textbf{0.568} ± 0.003 & 0.580 ± 0.022 \\
& U-Net      & 0.355 ± 0.026 & 0.391 ± 0.075 & 0.408 ± 0.032 & 0.457 ± 0.105 & \textbf{0.593} ± 0.005\\
\midrule
\multirow{3}{*}{200K}
& MetaSeg    & 0.678 ± 0.005 & 0.637 ± 0.017 & 0.563 ± 0.011 & 0.449 ± 0.054 & 0.454 ± 0.041 \\
& HierINRSeg & 0.738 ± 0.007 & 0.742 ± 0.005 & 0.696 ± 0.006 & 0.634 ± 0.008 & 0.592 ± 0.016 \\
& U-Net      & \textbf{0.781} ± 0.022 & \textbf{0.772 }± 0.041 & \textbf{0.790 } ± 0.001 & \textbf{0.774} ± 0.015 & \textbf{0.797} ± 0.029 \\
\bottomrule
\end{tabular}
\end{table}

\begin{table}[bt]
\centering
\caption{
Effect of adding augmentation to training at the 120K parameter budget. Each entry reports the Dice score for the model trained with augmentation, followed by the augmentation gain (as compared to the same model trained without augmentation).
}
\label{tab:augmentation_effect_120k}
\small
\setlength{\tabcolsep}{6pt}
\begin{tabular}{lccc}
\toprule
Dataset & MetaSeg & HierINRSeg & U-Net \\
\midrule
KKI (ID)       
& 0.688 (+0.019) & 0.750 (+0.001) & 0.701 (+0.016) \\
Caltech (OOD) 
& 0.569 (+0.040) & 0.676 (+0.013) & 0.654 (+0.100) \\
CMU (OOD)     
& 0.658 (+0.038) & 0.735 (+0.016) & 0.661 (+0.086) \\
\bottomrule
\end{tabular}
\end{table}

\textbf{Inference time.} For the implementation presented in the paper, HierINRSeg-10K takes 0.14s per 160×160 image on one RTX A6000. However, that uses the same 100-step inference fitting as MetaSeg for fair accuracy comparisons, which can be easily reduced substantially: for example, using 10 steps gives 0.018s per image with only a slight Dice score reduction of 0.03 averaged across three domains. Although slower than UNet-10K (0.003s per image with the same batch size=1 and 0.0009s with batch size=8), HierINRSeg-10K combines low absolute inference latency with substantially superior segmentation performance, offering a strong practical accuracy–efficiency trade-off, outperforming UNet-10K by 0.23 Dice at 100 fitting steps and 0.19 at 10 fitting steps. MetaSeg-10k has a similar speed as HierINRSeg, taking 0.13s per image for 100 fitting steps. When reduced to 10 steps, the time decreases to 0.014s while suffering a 0.02 Dice score reduction averaged across three domains.  

\textbf{Replacing the input coordinates.} Since the visualizations in Figures~\ref{fig:motiv0} and ~\ref{fig:motiv1} that motivated HierINRSeg show that features from earlier INR layers have the potential of conveying spatial information, we test an alternative design where we replace the input coordinates with features from the earliest selected INR layer, while the remaining INR features are integrated in the same way. The layer-wise ablations are repeated in Table~\ref{tab:emb_ablation}. The effect of this substitution is noticeable when only the latter INR layers are used, but diminishes as additional layers are included and becomes negligible for the full model, differing by at most 0.004 Dice across all datasets. This is consistent with our observations in Figures~\ref{fig:motiv0} and ~\ref{fig:motiv1} that earlier INR features retain spatial information, and further supports that the gains of HierINRSeg arise from integrating complementary features across the INR hierarchy.

\begin{table}[tbh]
\centering
\caption{
An alternative HierINRSeg variant where we replace the input coordinates to the segmentation head with the earliest selected INR layer. The numbers shown are the differences in Dice (embedded coordinates $-$ earliest-INR feature)
}
\label{tab:emb_ablation}
\small
\setlength{\tabcolsep}{4pt}
\begin{tabular}{llccccccc}
\toprule
\multirow{2}{*}{Selected INR layers} & \multicolumn{5}{c}{Selected layers} & \multicolumn{3}{c}{$\Delta$ Dice} \\
\cmidrule(lr){2-6} \cmidrule(lr){7-9}
 & 1 & 2 & 3 & 4 & 5 & KKI & Caltech & CMU \\
\midrule
1 layer & &  &  &  & \checkmark & +0.060 & +0.046 & +0.069 \\
2 layers & & &  & \checkmark & \checkmark & +0.039 & +0.043 & +0.018 \\
3 layers & &  & \checkmark & \checkmark & \checkmark & +0.002 & -0.011 & -0.004 \\
excl. middle &  \checkmark &  &  &  & \checkmark &+0.008  & +0.008 & +0.005 \\
4 layers &  & \checkmark & \checkmark & \checkmark & \checkmark & -0.004 & -0.019 & -0.007 \\
5 layers (all) & \checkmark & \checkmark & \checkmark & \checkmark & \checkmark & -0.004 & -0.002 & -0.003 \\
\bottomrule
\end{tabular}
\end{table}

\end{document}